%% file: main.tex
\documentclass{article} 
\usepackage{iclr2021_conference,times}

\input{math_commands.tex}

\usepackage[pdftex]{graphicx} 
\usepackage{comment}
\usepackage{hyperref}
\usepackage{url}
\usepackage{amsmath}
\usepackage{enumitem}
\usepackage{multirow}
\usepackage{booktabs}
\usepackage{wrapfig}
\usepackage{capt-of}
\usepackage{tikz}
\usepackage{pgfplots}
\usepackage{array}
\usepackage{xcolor,soul}
\usepackage{mathtools}
\usepackage{amsmath}

\usepackage{xcolor}

\title{Contrastive learning with Adversarial Perturbations for Conditional Text Generation}

\author{%
  Seanie Lee$^{1*}$,\quad Dong Bok Lee$^{1}$\thanks{Equal contribution},\quad Sung Ju Hwang$^{1,2}$\\
    KAIST$^{1}$,  AITRICS$^{2}$, South Korea \\
  \texttt{\{lsnfamily02, markhi, sjhwang82\}@kaist.ac.kr} \\  
}

\DeclareRobustCommand{\hlyellow}[1]{{\sethlcolor{yellow}\hl{#1}}}

\iclrfinalcopy 
\begin{document}

\maketitle

\input{0abstract}
\input{1introduction}
\input{2relatedwork}

\input{3method}
\input{4experiment}

\input{5conclusion}

\bibliography{iclr2021_conference}
\bibliographystyle{iclr2021_conference}

\newpage
\input{6appendix}

\end{document}

%% file: math_commands.tex
\usepackage{amsmath,amsfonts,bm}

\def\eqref#1{equation~\ref{#1}}

\def\1{\bm{1}}

\def\eps{{\epsilon}}

\def\ru{{\textnormal{u}}}
\def\rv{{\textnormal{v}}}

\def\rva{{\mathbf{a}}}
\def\rvb{{\mathbf{b}}}

\def\rvf{{\mathbf{f}}}
\def\rvg{{\mathbf{g}}}
\def\rvh{{\mathbf{h}}}
\def\rvu{{\mathbf{i}}}

\def\rvm{{\mathbf{m}}}

\def\rvu{{\mathbf{u}}}

\def\rvx{{\mathbf{x}}}
\def\rvy{{\mathbf{y}}}
\def\rvz{{\mathbf{z}}}

\def\rmH{{\mathbf{H}}}

\def\rmW{{\mathbf{W}}}

\DeclareMathAlphabet{\mathsfit}{\encodingdefault}{\sfdefault}{m}{sl}
\SetMathAlphabet{\mathsfit}{bold}{\encodingdefault}{\sfdefault}{bx}{n}

\DeclareMathOperator*{\argmin}{arg\,min}

%% file: 0abstract.tex
\begin{abstract}
Recently, sequence-to-sequence (seq2seq) models with the Transformer architecture have achieved remarkable performance on various conditional text generation tasks, such as machine translation. However, most of them are trained with \emph{teacher forcing} with the ground truth label given at each time step, without being exposed to incorrectly generated tokens during training, which hurts its generalization to unseen inputs, that is known as the ``exposure bias" problem. In this work, we propose to mitigate the conditional text generation problem by contrasting positive pairs with negative pairs, such that the model is exposed to various valid or incorrect perturbations of the inputs, for improved generalization. However, training the model with naïve contrastive learning framework using random non-target sequences as negative examples is suboptimal, since they are easily distinguishable from the correct output, especially so with models pretrained with large text corpora. Also, generating positive examples requires domain-specific augmentation heuristics which may not generalize over diverse domains. To tackle this problem, we propose a principled method to generate positive and negative samples for contrastive learning of seq2seq models. Specifically, we generate negative examples by adding small perturbations to the input sequence to minimize its conditional likelihood, and positive examples by adding  large perturbations while enforcing it to have a high conditional likelihood. Such ``hard'' positive and negative pairs generated using our method guides the model to better distinguish correct outputs from incorrect ones. We empirically show that our proposed method significantly improves the generalization of the seq2seq on three text generation tasks --- machine translation, text summarization, and question generation.
\end{abstract}

%% file: 1introduction.tex
\section{Introduction}
The sequence-to-sequence (seq2seq) models~\citep{seq2seq}, which learn to map an arbitrary-length input sequence to another arbitrary-length output sequence, have successfully tackled a wide range of language generation tasks.
Early seq2seq models have used recurrent neural networks to encode and decode sequences, leveraging attention mechanism \citep{attention} that allows the decoder to attend to a specific token in the input sequence to capture long-term dependencies between the source and target sequences. Recently, the Transformer~\citep{transformer}, which is an all-attention model that effectively captures long-term relationships between tokens in the input sequence as well as across input and output sequences, has become the de facto standard for most of the text generation tasks due to its impressive performance. Moreover, Transformer-based language models trained on large text corpora \citep{unilm, t5, bart} have shown to significantly improve the model performance on text generation tasks. 

\begin{figure}
	\begin{center}
		\includegraphics[width=1.0\linewidth]{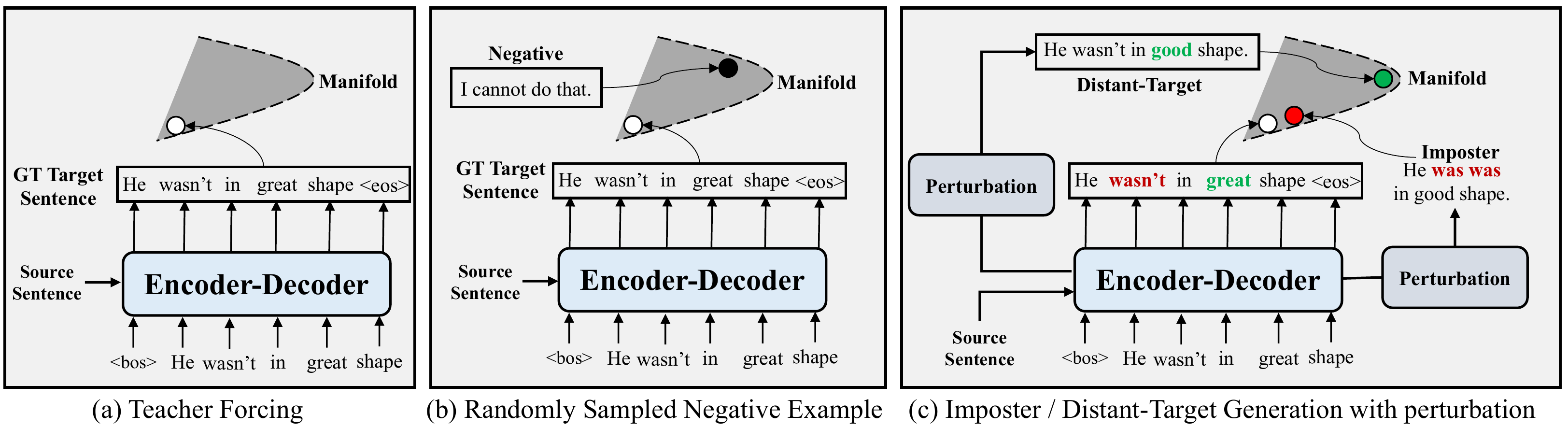}
	\end{center}
	\vspace{-0.20in}
	\caption{\small \textbf{Concept.} (a) Training seq2seq with \emph{teacher forcing}. (b) Naïve contrastive learning with randomly sampled negative examples. (c) Our method, CLAPS, which generates hard negative and positive examples. }\label{fig1}
\vspace{-0.28in}
\end{figure}

However, a crucial limitation of seq2seq models is that they are mostly trained only with \emph{teacher forcing}, where ground truth is provided at each time step and thus  never exposed to incorrectly generated tokens during training (Fig.~\ref{fig1}-(a)), which hurts its generalization. This problem is known as the ``exposure bias" problem \citep{exposure-bias} and often results in the generation of low-quality texts on unseen inputs. Several prior works tackle the problem, such as using reinforcement learning (RL) to maximize non-differentiable reward \citep{rl-mt, rl-sum}.

Another approach is to use RL or gumbel softmax \citep{gumbelsoftmax1} to match the distribution of generated sentences to that of the ground truth, in which case the reward is the discriminator output from a Generative Adversarial Network (GAN) \citep{gan-dialogue, feature-gan, seqgan}. Although the aforementioned approaches improve the performance of the seq2seq models on text generation tasks, they either require a vast amount of effort in tuning hyperparameters or stabilize training. 

\input{plot}

In this work, we propose to mitigate the exposure bias problem with a simple yet effective approach, in which we \emph{contrast} a positive pair of input and output sequence to negative pairs, to expose the model to various valid or incorrect sentences. Naïvely, we can construct negative pairs by simply using random non-target sequences from the batch~\citep{simclr}. However, such a naïve construction yields meaningless negative examples that are already well-discriminated in the embedding space (Fig.~\ref{fig1}-(b)), which we highlight as the reason why existing methods~\citep{simclr} require large batch size. This is clearly shown in Fig.~\ref{cont-batch}, where a large portion of positive-negative pairs can be easily discriminated without any training, which gets worse as the batch size decreases as it will reduce the chance to have meaningfully difficult examples in the batch. Moreover, discriminating positive and naïve negative pairs becomes even more easier for models pretrained on large text corpora. 

To resolve this issue, we propose principled approaches to automatically generate negative and positive pairs for constrastive learning, which we refer to as \emph{Contrastive Learning with Adversarial Perturbation for Seq2seq learning} (CLAPS). Specifically, we generate a negative example by adding a small perturbation to the hidden representation of the target sequence, such that its conditional likelihood is minimized (Denoted as the red circle in Fig.~\ref{fig1}-(c)). Conversely, we construct an additional positive example (Denoted as green circle in Fig.~\ref{fig1}-(c)) by adding a large amount of perturbation to the hidden representation of target sequence such that the perturbed sample is far away from the source sequence in the embedding space, while enforcing it to have high conditional likelihood by minimizing Kullback-Leibler (KL) divergence between the original conditional distribution and perturbed conditional distribution. This will yield a negative example that is very close to the original representation of target sequence in the embedding space but is largely dissimilar in the semantics, while the generated positive example is far away from the original input sequence but has the same semantic as the target sequence. This will generate difficult examples that the model fails to correctly discriminate (Fig.~\ref{fig1}-(c), Fig.2), helping it learn with more meaningful pairs.

To verify the efficacy of our method, we empirically show that it significantly improves the performance of seq2seq model on three conditional text generation tasks, namely machine translation, text summarization and question generation. Our contribution in this work is threefold:

\vspace{-0.12in}
\begin{itemize}[itemsep=1mm, parsep=0pt, leftmargin=*]
    \item To mitigate the exposure bias problem, we propose a contrastive learning framework for conditional sequence generation, which contrasts a positive pair of source and target sentence to negative pairs in the latent embedding space, to expose the model to various valid or incorrect outputs.
    
    \item To tackle the ineffectiveness of conventional approach for constructing negative and positive examples for contrastive learning, we propose a principled method to automatically generate negative and positive pairs, that are more difficult and allows to learn more meaningful representations.
    
    \item We show that our proposed  method, CLAPS, significantly improves the performance of seq2seq model on three different tasks: machine translation, text summarization, and question generation.
\end{itemize}
\vspace{-0.1in}

%% file: plot.tex
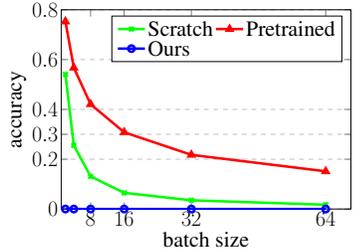
\begin{wrapfigure}{t}{0.33\textwidth}
	\begin{center}
	\vspace{-0.35in}
	\captionof{figure}{\small Accuracy of classifying a positive pair from negative pairs varying batch size \textbf{without any training.}}
    \label{cont-batch}
    \begin{tikzpicture}[thick,scale=0.6, every node/.style={scale=0.55}]
		\begin{axis}[
		compat=1.10,
		xlabel={batch size},
		ylabel={accuracy},
		legend cell align={left},
		legend columns=2,
		xmin=1, xmax=70,
		ymin=0, ymax=0.8,
		xtick={8,16,32,64},
		ytick={1.0, 0.8,0.6, 0.4, 0.3, 0.2, 0.0 },
		legend pos= north east,
		ymajorgrids=true,
		grid style=dashed,
		legend style={font=\Huge},
		label style={font=\Huge},
		tick label style={font=\Huge},
        width=8cm, height=6cm,
        every axis plot/.append style={ultra thick}
		]
		\addplot[
		color=green,
		mark=x,
		]
		coordinates {
			(2,0.5409)(4,0.2550)(8,0.1309)(16, 0.0649)(32, 0.0351)(64, 0.0171)
		};
		\addplot[
		color=red,
		mark=triangle,
		]
		coordinates {
			(2, 0.7530)(4, 0.5675)(8,0.4208)(16, 0.3082)(32, 0.2176)(64, 0.1515)
		};
		\addplot[
		color=blue,
		mark=o,
		]
		coordinates {
			(2, 0.0)(4, 0.0)(8, 0.0)(16,0.0)(32,0.0)(64,0.0)
		};
		\legend{Scratch, Pretrained, Ours}
		\end{axis}
		\end{tikzpicture}
	\end{center}
	\vspace{-0.295in}
\end{wrapfigure}

%% file: 2relatedwork.tex
\section{Related Work}

\noindent
\textbf{Exposure Bias}
There are several prior works to tackle the exposure bias \citep{exposure-bias}. \cite{scheduled-sampling} introduce scheduled sampling where the model is initially guided with the true previous tokens but uses the tokens generated by the seq2seq model as the conditional input for the next token, as training goes on.
\cite{rl-sum, rl-mt} leverage RL to maximize non-differentiable rewards, so it enables to penalize the model for incorrectly generated sentences. Another works \citep{feature-gan, gan-dialogue, seqgan} train GANs to match the distribution of generated sequences to that of ground truth. Since sampling tokens from the generator is not differentiable, they resort RL or gumbel-softmax to train the networks in end-to-end fashion. However, they require either a large amount of effort to tune hyperparameters or stabilize training. However, \cite{rl-not-good} show that RL for machine translation does not optimize the expected reward and the performance gain is attributed to the unrelated effects such as increasing the peakiness of the output distribution. Moreover,  \citep{gan-falling-short} show that by tuning the temperature parameter, the language models trained with MLE can be tuned to  outperform GAN-based text generation models.

\noindent
\textbf{Adversarial Perturbation} Many existing works, such as \citep{pgd}, address the robustness of neural networks to adversarial examples, which are generated by applying a small perturbations to the input samples. While adversarial robustness has been mostly explored in image domains, \cite{virtual} adopted adversarial training to text domains. However instead of targeting robustness to perturbed samples, they utilize the adversarial examples as augmented data, and enforce consistency across the predictions across original unlabeled example and its perturbation, for semi-supervised learning. Recently \cite{freelb} and \cite{smart} leverage adversarial training to induce the smoothness of text classifiers, to prevent overfitting to training samples. While they are relevant to ours, these methods do not have the notion of positive and negative examples as they do not consider contrastive learning, and only target text classification. Moreover, they are computationally prohibitive since they use PGD for adversarial training, which requires iterative optimization for each individual sample. Recently, \cite{better-finetuning} propose a simpler yet effective method based on Gaussian noise perturbation to regularize neural networks without expensive PGD steps, which is shown to outperform methods from~\cite{freelb} and~\cite{smart}. Although our work is similar to these prior works in that we add perturbations to the text embeddings, note that we used the adversarially-generated samples as negative examples of our contrastive learning framework rather than trying to learn the model to be robust to them.

\noindent
\textbf{Contrastive Learning}
Contrastive learning has been widely used. It is to learn a representation by contrasting positive pairs and negative pairs.  \citet{triplet1, lmnn, triplet3} leverage a triplet loss to separate positive examples from negative examples in metric learning. \citet{simclr} show that contrastive learning can boost the performance of self-supervised and semi-supervised learning in computer vison tasks. In natural language processing (NLP), contrastive learning has been widely used. In Word2Vec \citep{w2v}, neighbouring words are predicted from context with noise-contrastive estimation \citep{nce}. Beyond word representation, \cite{quick-thought} sample two contiguous sentences for positive pairs and the sentences from other document as negative pairs. They constrast positive and negative pairs to learn sentence representation. Moreover, contrastive learning has been investigated in various NLP tasks --- language modeling \citep{margin-lm}, unsupervised word alignment \citep{alignment}, caption generation \citep{cap-gen1, cap-gen2}, and machine translation \citep{omission-cont}.

%% file: 3method.tex
\section{Method}
\subsection{Background: Conditional text Generation}
The goal of conditional text generation with a seq2seq model is to generate an output text sequence $\rvy^{(i)} = (y^{(i)}_1, \ldots, y^{(i)}_T)$ with length $T$ conditioned on the input text sequence $\rvx^{(i)} = (x^{(i)}_1, \ldots, x^{(i)}_L)$ with length $L$. A typical approach to the conditional text generation is to leverage the encoder-decoder architecture to parameterize the conditional distribution. We maximize the conditional log likelihood $\log p_\theta(\rvy| \rvx)$ for a given $N$ observations $\{(\rvx^{(i)}, \rvy^{(i)})\}_{i=1}^N$ as follows:
\begin{align}
\begin{split}
    \mathcal{L}_{MLE}(\theta) &= \sum_{i=1}^N\log p_\theta (\rvy^{(i)}|\rvx^{(i)}) \\
    p_\theta(y^{(i)}_1, \ldots, y^{(i)}_T | \rvx^{(i)}) &= \prod_{t=1}^T p_\theta(y^{(i)}_t |\rvy^{(i)}_{<t}, \rvx^{(i)}) \\
    p_\theta(y^{(i)}_t | \rvy^{(i)}_{<t}, \rvx^{(i)}) &= \text{softmax}(\rmW\rvh^{(i)}_t + \rvb) \\
    \rvh^{(i)}_t &= g(y^{(i)}_{t-1}, \mathbf{M}^{(i)}; \theta), \:\: \mathbf{M}^{(i)} = f(\rvx^{(i)} ; \theta)\\
\end{split}
\label{nll}
\end{align}
where $f, g$ denote the encoder and the decoder respectively and $\mathbf{M}^{(i)} =[\rvm^{(i)}_1 \cdots \rvm^{(i)}_L] \in \mathbb{R}^{d\times L}$ is the concatenation of the hidden representations of the source tokens $\rvx^{(i)}$. 

\subsection{Contrastive Learning with Adversarial Perturbations for Seq2Seq}
Since most of the seq2seq models are trained with teacher forcing where the ground truth tokens are provided to maximize Eq.~\ref{nll}, they are never exposed to incorrectly generated tokens during training, which is known as the ``expousre bias" problem. In order to tackle the problem, we propose a contrastive learning framework to expose the model to various valid or incorrect output sequences for a given input sentence. Following the contrastive learning framework~\citep{simclr}, we can train the model to learn the representations of the ground truth sentence by contrasting the positive pairs with the negative pairs, where we select the negative pairs as a random non-target output sequence from the same batch. As shown in Fig.~\ref{method-fig}-(a), we project the source and target text sequences onto the latent embedding space. Then we maximize the similarity between the pair of source and target sequence, while minimizing the similarity between the negative pairs as follows:
\begin{align}
    \begin{split}
        \mathcal{L}_{cont}(\theta) &=  \sum_{i=1}^N \log \frac{\exp({\text{sim}(\rvz^{(i)}_\rvx, \rvz^{(i)}_\rvy)/\tau})}{\sum_{\rvz^{^{(j)}}_\rvy \in S} \exp(\text{sim}(\rvz^{(i)}_\rvx, \rvz^{^{(j)}}_\rvy)/\tau)} \\
        \rvz^{(i)}_\rvx &=  \xi(\mathbf{M}^{(i)};\theta), \: \rvz^{(i)}_\rvy = \xi(\mathbf{H}^{(i)}; \theta) \\
        \xi([\rv_1 \cdots \rv_T]; \theta) &\coloneqq \text{AvgPool}([\rvu_1 \cdots \rvu_T]) \text{, where } \ru_t =  \text{ReLU}(\mathbf{W}^{(1)}\rv_t + \mathbf{b}^{(1)}) 
    \label{naive-cont}
    \end{split}
\end{align}

where $\xi$ denotes the composition of affine transformation with the ReLU \citep{relu} and average pooling to compute the fixed sized representation of a sentence $\rvz \in \mathbb{R}^d$, $\mathbf{H}^{(i)} =[\mathbf{h}^{(i)}_1 \cdots \mathbf{h}^{(i)}_T ] \in \mathbb{R}^{d \times T}$ is a concatenation of the decoder hidden states of  the target sentence $\rvy^{(i)}$ across all the time steps. Furthermore, $S=\{\rvz^{(j)}_{\rvy}: j \neq i \}$ is a set of hidden representations of target sentences (the objects other than circles in Fig.~\ref{method-fig}-(a)) that are  randomly sampled and not paired with the source sentence $\rvx^{(i)}$, and sim$(\cdot, \cdot)$ is a cosine similarity function. 

However, training the model with naïve contrastive learning framework using random non-target sequences as negative examples is highly suboptimal, as described in the introduction and shown in Fig.~\ref{fig1}. Many of such naïve negative examples are often located far away from the positive examples in the embedding space from the beginning, when using the pretrained language model. Therefore, simply using the examples from the same batch as done in~\cite{simclr} will result in trivial negative examples and require very large batch size to enable sampling meaningful negative pairs within the same batch. Moreover, generating positive examples for text sequences is not a trivial problem either since for text domains, we do not have a well-defined set of augmentation methods that preserves the input semantics, unlike with the image domains. To tackle such difficulties, we propose a principled method to automatically construct the adversarial negative and positive examples, such that the samples are difficult for the model to classify correctly. These adversarial positive/negative pairs can guide the model to learn a more accurate representation of the target text sequence, by identifying which features make the output positive or negative (See Fig.~\ref{fig1}-(c)). 

\subsection{Generation of Imposters}
\begin{figure}
	\begin{center}
		\includegraphics[width=1.0\linewidth]{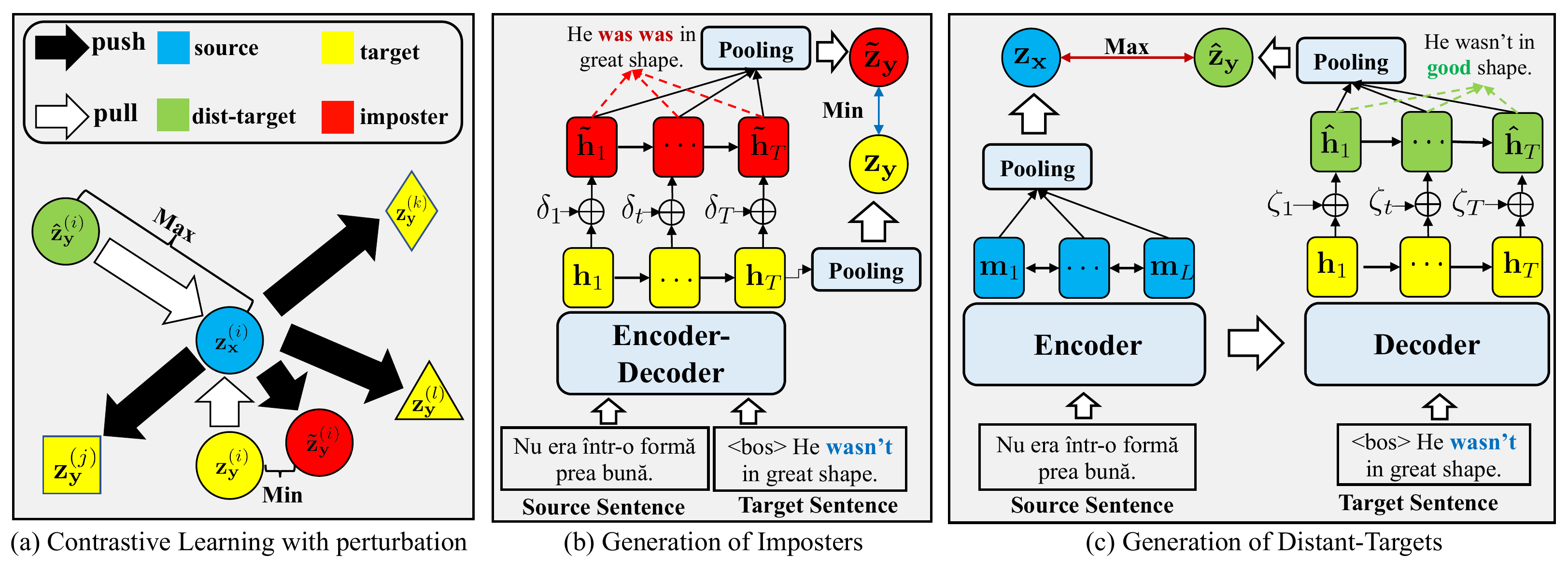}
	\end{center}
	\vspace{-0.22in}
	\caption{\small \textbf{Generation of imposters and distant-targets with perturbation. } (a) We add small perturbation $\delta_t$ to $\rvh_t$ for $\Tilde{\rvz}_\rvy $ so that its conditional likelihood is minimized to generate an invalid sentence. (b) We add large perturbation $\zeta_t$ to $\rvh_t$ for $\hat{\rvz}_\rvy$ by maximizing the distance from $\rvz_\rvx$, the representation of source sentence but enforcing its likelihood high to preserve the original semantics. }\label{method-fig}
\vspace{-0.15in}
\end{figure}

As shown in Fig.~\ref{method-fig}-(b), to generate a negative example, we add a small perturbation $\boldsymbol{\delta}^{(i)} = [\delta^{(i)}_1 \cdots  \delta^{(i)}_T] \in \mathbb{R}^{d\times T}$ to the $\mathbf{H}^{(i)}$, which is the hidden representation of target sequence $\rvy^{(i)}$, such that its conditional likelihood is minimized as follows:
\begin{align}
\begin{split}
     \mathbf{\Tilde{H}}^{(i)} = \mathbf{H}^{(i)} + \boldsymbol{\delta}^{(i)} \text{ where } \boldsymbol{\delta}^{(i)} &= \argmin_{\boldsymbol{\delta}, ||\boldsymbol{\delta}||_2 \leq \epsilon} \log p_\theta (\rvy^{(i)} | \rvx^{(i)}; \mathbf{H}^{(i)}+\boldsymbol{\delta}) \\
      p_\theta (\rvy^{(i)} | \rvx^{(i)}; \mathbf{H}^{(i)}+\boldsymbol{\delta}) &= \prod_{t=1}^T p_\theta(y^{(i)}_t|\rvy^{(i)}_{<t}, \rvx^{(i)}; \rvh^{(i)}_t + \delta_t) \\
      p_\theta (y^{(i)}_t|\rvy^{(i)}_{<t}, \rvx^{(i)};\rvh^{(i)}_t + \delta_t) &= \text{softmax}\{\rmW(\rvh^{(i)}_t + \delta_t) + \rvb\} \text{, where } \delta_t \in \mathbb{R}^{d}
\end{split}
\label{neg-pert}
\end{align}
The exact minimization of the conditional log likelihood with respect to $\boldsymbol{\delta}$ is intractable for deep neural networks. Following \cite{adv-example}, we approximate it by linearizing $\log p_\theta (\rvy^{(i)}|\rvx^{(i)})$ around $\mathbf{H}^{(i)}$ as follows:
\begin{align}
    \mathbf{\Tilde{H}}^{(i)} = \mathbf{H}^{(i)} - \epsilon \frac{\boldsymbol{g}}{|| \boldsymbol{g}||_2} \text{, where } \boldsymbol{g} = \nabla_{\mathbf{H}^{(i)}} \log p_\theta (\rvy^{(i)} | \rvx^{(i)})
\end{align}

We add small perturbation to the hidden representation of each token of target sentence $\rvy^{(i)}$ such that its conditional likelihood is minimized. Thus, the perturbed $\mathbf{\Tilde{H}}^{(i)}$, which we call an \emph{imposter} (inspired by~\cite{lmnn}), is semantically very dissimilar to $\rvy^{(i)}$, but very close to the hidden representation $\rmH^{(i)}$ in the embedding space (Fig.~\ref{method-fig}-(a)). This will make it non-trivial for the sequence-to-sequence model to distinguish it from the representation of true target sequence $\rvy^{(i)}$. Please note while adversarial perturbations are generated similarly as in~\cite{virtual}, we use them in a completely different way. While they train the model to be invariant to adversarial samples within the $\epsilon$-ball, we push them far away from the source sentence while pulling the ground truth target sentence to the input sentence. In other words, we use the perturbed representation as an additional negative sample for contrastive learning as follows:
\begin{equation}
        \mathcal{L}_{cont-neg}(\theta) =  \sum_{i=1}^N\log \frac{\exp({\text{sim}(\rvz^{(i)}_\rvx, \rvz^{(i)}_\rvy)/\tau})}{\sum_{\rvz^{(k)}_\rvy \in S\cup \{\Tilde{\rvz}^{(i)}_{\rvy} \}} \exp(\text{sim}(\rvz^{(i)}_\rvx, \rvz^{(k)}_\rvy)/\tau)}  \text{, where } \Tilde{\rvz}^{(i)}_{\rvy} =  \xi(\Tilde{\mathbf{H}}^{(i)}; \theta)
    \label{adv-cont}
\end{equation}
Alternatively, we can generate an imposter by perturbing the hidden representation of target sentence $\rvy$ so that its conditional likelihood is minimized but very close to the source sentence $\rvx$ in the embedding space. However, we empirically find that such a variation yields less performance gain.

\subsection{Generation of Distant-Targets}
Moreover, as shown in Fig.~\ref{method-fig}-(c), we construct an additional positive pair of source sequence $\rvx^{(i)}$ by adding large perturbation $\boldsymbol{\zeta}^{(i)} = [\zeta^{(i)}_1 \cdots \zeta^{(i)}_T] \in \mathbb{R}^{d \times T}$ to $\mathbf{H}^{(i)}$ the hidden state of target sequence $\rvy^{(i)}$, such that cosine similarity from $\rvz^{(i)}_\rvx$ is minimized, but the conditional likelihood is enforced to remain high. However, the exact computation of $\boldsymbol{\zeta}^{(i)}$ with such constraints is intractable. We approximate it with the following two separate stages.  First, we add perturbation to $\mathbf{H}^{(i)}$ such that it minimizes the contrastive learning objective $\mathcal{L}_{cont}(\theta)$ as shown in Eq.~\ref{pos-pert1}. Then we add another perturbation to minimize the KL divergence between perturbed conditional distribution $p_\theta(\hat{y}^{(i)}_t | \hat{\rvy}^{(i)}_{<t}, \rvx^{(i)})$ and the original conditional distribution $p_\theta(y^{(i)}_t|\rvy_{<t}^{(i)}, \rvx^{(i)})$ as shown in Eq.~\ref{pos-pert2}, where $\mathbf{\overline{H}} = [\overline{\mathbf{h}}_1 \cdots \overline{\mathbf{h}}_T] \in \mathbb{R}^{d \times T}$, $\mathbf{\hat{H}} = [\hat{\mathbf{h}}_1 \cdots \hat{\mathbf{h}}_T] \in \mathbb{R}^{d \times T}$, and $\eta \in \mathbb{R}$. Note that $\theta^{*}$ denotes the copied of the model parameter $\theta$ and is considered to be constant to prevent it from being updated through back-propagation.
\begin{align}
\overline{\mathbf{H}}^{(i)} &= \mathbf{H}^{(i)} - \eta \frac{\rvg}{||\rvg||_2} \text{ where } \rvg = \nabla_{\mathbf{H}^{(i)}} \mathcal{L}_{cont}(\theta)
\label{pos-pert1}
\end{align}
\vspace{-0.15in}
\begin{align}
\begin{split}
    p_\theta(\hat{y}^{(i)}_t | \hat{\rvy}^{(i)}_{<t}, \rvx^{(i)}) &= \text{softmax}(\rmW\overline{\mathbf{h}}^{(i)}_t + \mathbf{b}) \\
\mathcal{L}_{KL}(\theta) &= \sum_{i=1}^N\sum_{t=1}^T D_{KL}(p_{\theta^{*}}(y^{(i)}_t| \rvy^{(i)}_{<t},\rvx^{(i)}) || p_\theta(\hat{y}^{(i)}_t | \hat{\mathbf{y}}^{(i)}_{<t}, \rvx^{(i)}) \\
\hat{\mathbf{H}}^{(i)} &= \overline{\mathbf{H}}^{(i)} - \eta \frac{\rvf}{||\rvf||_2} 
\text{, where } \rvf =\nabla_{\overline{\mathbf{H}}_1^{(i)}} \mathcal{L}_{KL}(\theta) 
\label{pos-pert2}
\end{split}
\end{align}
We consider the perturbed hidden state $\hat{\mathbf{H}}^{(i)}$ as an additional positive example for source sequence $\rvx^{(i)}$, which we refer to as a \emph{distant-target}. We can use a distant-target to augment contrastive learning and minimize $\mathcal{L}_{KL}(\theta)$ as follows:
\begin{align}
        \mathcal{L}_{cont-pos}(\theta) =  \sum_{i=1}^N\log \frac{\exp({\text{sim}(\rvz^{(i)}_\rvx, \hat{\rvz}^{(i)}_\rvy)/\tau})}{\sum_{\rvz^{(k)}_\rvy \in S\cup \{\Tilde{\rvz}^{(i)}_{\rvy} \}} \exp(\text{sim}(\rvz^{(i)}_\rvx, \rvz^{(k)}_\rvy)/\tau)} \text{, where } \hat{\rvz}^{(i)}_\rvy = \xi(\hat{\mathbf{H}}^{(i)}; \theta)
    \label{pos-cont}
\end{align}

\paragraph{CLAPS objective}
Incorporating the loss on the imposter and the distant target introduced above, we estimate the parameters of the seq2seq model $\theta$ by maximizing the following objective, where $\alpha, \beta$ are hyperparameters which control the importance of contrastive learning and KL divergence:
\begin{equation}
 \max_\theta \mathcal{L}_{MLE}(\theta) - \alpha \mathcal{L}_{KL}(\theta) + \beta \{\mathcal{L}_{cont-neg}(\theta) + \mathcal{L}_{cont-pos}(\theta)\}
 \label{objective}
\end{equation}
For all the experiments, we set $\alpha$ and $\beta$ as 1, which we search through cross-validation. Note that after training is done, we remove the pooling layer $\xi$ and generate text with the decoder $g$, given an input encoded with the encoder $f$.

%% file: 4experiment.tex
\section{Experiment}
We validate our method on benchmark datasets on three conditional text generation tasks. 
\subsection{Tasks}
\textbf{Machine Translation (MT)}
For machine translation, we use WMT16 Romanian-English parallel corpus (WMT'16 RO-EN) to train the model. We tokenize the pairs of source and target sequences with the same tokenizer as \cite{t5}. We finetune the pretrained T5-small model for 20 epochs with the batch size of 128 and Adafactor \citep{adafactor}. For contrastive learning, we set the norm of perturbation, $\eta$ and $\epsilon$ as 3.0.

\noindent
\textbf{Text Summarization (Sum.)} 
For text summarization, we use XSum dataset \citep{xsum} of which summaries are highly abstractive, thus extractive summarization models under-perform abstractive models. We follow the most of the experimental settings for machine translation as described above, except that we set the norm of perturbation, $\eta$ and $\epsilon$ as 1.0 and 1.0, respectively.

\noindent
\textbf{Question Generation (QG)}
For question generation, we aim to generate a question from a given answer and paragraph, i.e., we model conditional distribution $p_\theta(\rvy| \rvx, \rva)$ where $\rvx, \rvy, \rva$ denote a paragraph, question and answer, respectively.  We concatenate the answer and paragraph with special tokens to generate the question conditioned on both of the answer and paragraph. As the previous experimental settings, we finetune T5-small model on SQuAD dataset \citep{squad} for 20 epochs with batch size 128 and set the norm of  perturbation, $\eta$ as $3.0$ and $\eps$ as $1.0$. Since the test set of SQuAD is only accessible via leader board, we randomly split the validation set into a validation set and a test set.

\begin{table}
	\small
	\centering
	\resizebox{0.95\textwidth}{!}{
	\begin{tabular}{lccccccc}
		\midrule[0.8pt]
		{\textbf{Method}} & \textbf{Aug.} & {\textbf{BLEU-1}} & {\textbf{BLEU-2}} & {\textbf{BLEU-3}} & {\textbf{BLEU-4}} & {\textbf{BLEU}} & {\textbf{F1/EM}} \\
		\midrule[0.8pt]
		\multicolumn{8}{c}{ \textbf{Question Generation} - SQuAD} \\ 
		\midrule[0.8pt]
		{Harvesting-QG} & {-}& {-} & {-} & {20.90} & {15.16} & {-} & {66.05/54.62}\\
		{T5-MLE } & {-} &{41.26} & {30.30} & {23.38} & {18.54} & {21.00} & {67.64/55.91}\\
		{{$\alpha$-T5-MLE ($\alpha=0.7$) }} & {-} &{40.82} & {29.79} & {22.84} & {17.99} & {20.50} & {68.04/56.30}\\
		{{$\alpha$-T5-MLE ($\alpha=2.0$) }} & {-} & {37.35} & {27.20} & {20.79} & {16.36} & {18.41} & {65.74/54.76}\\
		{T5-SSMBA } & {Pos.} &{41.67} & {30.59}& {23.53} & {18.57} & {21.07} &{68.47/56.37}\\ 
		{T5-WordDropout Contrastive} & {Neg.} & {41.37} & {30.50} & {23.58} & {18.71} & {21.19} & {68.16/56.41}\\
		{{R3F }} & {-} & {41.00} & {30.15} & {23.26} & {18.44} & {20.97} & {65.84/54.10}\\
		{T5-MLE-contrastive} & {-} & {41.23} & {30.28} & {23.33} & {18.45} & {20.91} & {67.32/55.25}\\
		\midrule[0.5pt]
		{\bf T5-CLAPS w/o negative} & {Pos.} & {41.87} & {30.93} & {23.90} & {18.92} & {21.38} & {-}\\
		{\bf T5-CLAPS w/o positive} & {Neg.} & {41.65} & {30.69} & {23.71} & {18.81} & {21.25} &{68.26/56.41}\\
		{\bf T5-CLAPS} & {Pos.+Neg.} & \textbf{42.33} & \textbf{31.29} & \textbf{24.22} & \textbf{19.19} & \textbf{21.55} &\textbf{69.01/57.06}\\
		\midrule[0.5pt]
		{ERNIE-GEN \citep{ernie}}& {-} &{-} & {-} & {-} & \textbf{26.95} & {-} & {-} \\ 
		{Info-HCVAE \citep{info-hcvae}}& {-} &{-} & {-} & {-} & {-} & {-} &\textbf{81.51/71.18} \\ 
		\bottomrule[0.8pt]
		\multicolumn{8}{c}{\textbf{Machine Translation} - WMT'16 RO-EN } \\ 
		\midrule[0.8pt]
		{Transformer} & {-} & {50.36} & {37.18} & {28.42} & {22.21}  & {26.17}\\
		{Scratch-T5-MLE} & {-} &{51.62} & {37.22} & {27.26} & {21.13} & {25.34}\\
		{Scratch-CLAPS} & {Pos.+Neg.} & {53.42} & {39.57} & {30.24} & {23.59} & {27.61}\\
		{T5-MLE} & {-} &{57.76} & {44.45} & {35.12} & {28.21} & {32.43}\\
		{$\alpha$-T5-MLE ($\alpha=0.7$) } & {-} & {57.63} & {44.23} & {33.84} & {27.90} & {32.14} \\
		{$\alpha$-T5-MLE ($\alpha=2.0$)} & {-} & {56.03} & {42.59} & {33.29} & {26.45} & {30.72} \\
		{T5-SSMBA} & {Pos.} & {58.23} & {44.87} & {35.50} & {28.48} & {32.81}\\ 
		{T5-WordDropout Contrastive} & {Neg.}& {57.77} & {44.45} & {35.12} & {28.21} & {32.44} \\
		{R3F} & {-} & {58.07} & {44.86} & {35.57} & {28.66} & {32.99}\\
		{T5-MLE-contrastive} &{-}& {57.64} & {44.12} & {34.74} & {27.79} & {32.03} \\
		\midrule[0.5pt]
		{\bf T5-CLAPS w/o negative} & {Pos.} & {58.81} & {45.52} & {36.20} & {29.23} & {33.50} & {67.58/55.91} \\
		{\bf T5-CLAPS w/o positive} & {Neg.} & {57.90} & {44.60} & {35.27} & {28.34} & {32.55} \\
		{\bf T5-CLAPS} & {Pos.+Neg.} & \textbf{58.98} & \textbf{45.72} & \textbf{36.39} & \textbf{29.41} & \textbf{33.96} \\
		\midrule[0.5pt]
		{\cite{clm}} & {-} & {-} & {-} & {-} & {-} & {\textbf{38.5}} \\
		\bottomrule[0.8pt]
	\end{tabular}
	}
	\vspace{-0.1in}
	\caption{\small BLEU scores on WMT'16 RO-EN and SQuAD for machine translation and question generation. EM/F1 scores with BERT-base QA model for question generation.}
	\vspace{-0.25in}
	\label{mt-qg}
\end{table}

\subsection{Experimental Setups}
\textbf{Implementation Details} For the encoder $f$, and decoder $g$, we use T5-small model, which is based on transformer with the hidden dimension, $d=512$. We set the temperature, $\tau$ as 0.1 for all the experiments. At test time, we use beam search of width 4 to generate the target sequences.
\noindent
\textbf{Common Baselines} We compare our method against relevant baselines.
\vspace{-0.15in}
\begin{enumerate}[itemsep=0.8mm, parsep=0pt, leftmargin=*]
    \item \textbf{T5-MLE}: A pretrained T5 model fine-tuned to maximize $\mathcal{L}_{MLE}(\theta)$.
    
    \item \textbf{Scratch-T5-MLE}: A random initialized Transformer model that has the identical architecture to T5, trained by maximizing $\mathcal{L}_{MLE}(\theta)$.
    
    \item \textbf{$\alpha$-T5-MLE}: T5 model trained with MLE, with varying temperature $\alpha$ in the softmax function when decoding the target sentences, as done in~\cite{gan-falling-short}
    
    \item \textbf{T5-SSMBA}: This is the T5 model trained to maximize $\mathcal{L}_{MLE}(\theta)$, with additional examples generated by the technique proposed in~\cite{ssmba}. which are generated by corrupting the target sequences and reconstructs them using a masked language model, BERT.
    
    \item \textbf{T5-WordDropout Contrastive}: This is a T5 model trained with the contrastive learning framework proposed in ~\cite{omission-cont}, which heuristically generates negative examples by removing the most frequent word from the target sequence. We pretrain T5-small to maximize $\mathcal{L}_{MLE}(\theta)$ and further train the model to assign higher probability to the ground truth target sentence than a negative example with max-margin loss. 
    
    \item \textbf{R3F}: This is a T5 model that minimizes the negative log likelihood and symmetric KL-divergence between original conditional log likelihood $p_\theta(\rvy|\rvx)$ and $p_\theta(\rvy|\Tilde{\rvx}) $ to enforce the function to be smooth, $\text{where } \Tilde{\rvx} = \text{WordEmbedding}(\rvx) + \rvz, \rvz = (z_1, \ldots, z_L), z_i \stackrel{i.i.d}{\sim} \mathcal{N}(\mathbf{0}, \text{diag}(\sigma_1, \ldots, \sigma_d))$.

    \item \textbf{T5-MLE-contrastive}: This is a naive constrastive learning framework with positive/negative pairs, which maximizes the contrastive learning objective from Eq.~\ref{naive-cont}.
    
    \item \textbf{T5-CLAPS w/o positive (negative)}: Our proposed model which jointly maximizes the log likelihood and the contrastive learning objective with imposters but does not use any distant-targets or imposters.
    
    \item \textbf{T5-CLAPS}: Our full model which jointly maximizes the log likelihood, contrastive learning objective, and KL-divergence as described in the Eq.~\ref{objective}.
    
    \item \textbf{Scratch-CLAPS}: Our full model as \textbf{T5-CLAPS} but with randomly initialized T5 architecture.
\end{enumerate}

\textbf{Task specific baselines}
For machine translation, we use the Transformer \citep{transformer} which consists of 6 layers of self-attention layer with 8 multi-head attention and 512 dimension, as an additional baseline. For QG, we additionally compare our models against Harvesting-QG \citep{harvesting}, which is a LSTM model with copy mechanism. For text summarization, we use PTGEN-COVG \citep{pointer-generator} as a baseline, which uses copy mechanism and coverage to handle out of vocabulary word and prevent word repetition, and CONVS2S~\citep{xsum} which uses convolutional networks as the encoder and decoder.

\textbf{Evaluation Metric} Following the conventional evaluation metrics, we adopt n-gram BLEU and BLEU \citep{bleu} for MT and QG. For text summarization, we use Rouge \citep{rouge} and Meteor \citep{meteor}. As an additional performance measure for question generation, we evaluate a BERT QA model on the SQuAD test set, where the QA model is  trained with the questions generated by each QG methods from the contexts and answers of HarvestingQA dataset~\citep{harvesting}, and report the F1 and Exact Match (EM).

\begin{table}
	\small
	\centering
	\caption{\small Rouge and Meteor on Xsum test set for text summarization.}
	\begin{tabular}{lccccc}
		\midrule[0.8pt]
		{\textbf{Method}} & \textbf{Aug.}& {\textbf{Rouge-1}} & {\textbf{Rouge-2}} & {\textbf{Rouge-L}} & {\textbf{METEOR}}  \\
		\midrule[0.8pt]
		\multicolumn{6}{c}{\textbf{Text Summarization} - XSum } \\ 
		\midrule[0.8pt]
		{PTGEN-COVG} &  {-}& {28.10} & {8.02} & {21.72} & {12.46} \\
		{CONVS2S} & {-}& {31.89} & {11.54} & {25.75} & {13.20}  \\
		{Scratch-T5-MLE} & {-} & {31.44} & {11.07} & {25.18} & {13.01} \\
		{Stcratch-CLAPS} & {Pos.+Neg.} & {33.52} & {12.59}  & {26.91} & {14.18}\\
		{T5-MLE} & {-}&{36.10} & {14.72} & {29.16} & {15.78}\\
		{{$\alpha$-T5-MLE ($\alpha=0.7$) }} & {-} & {36.68} & {15.10} & {29.72} & {15.78} \\
		{{$\alpha$-T5-MLE ($\alpha=2.0$) }} & {-} & {34.18} & {13.53} & {27.35} & {14.51} \\
		{T5-SSMBA} & {Pos.}& {36.58} & {14.81} & {29.68} & {15.38} \\ 
		{T5-WordDropout Contrastive} & {Neg.}& {36.88} & {15.11} & {29.79} & {15.77} \\
        {R3F } & {-} & {36.96} & {15.12} & {29.76} & {15.68}\\
        {T5-MLE-contrastive} &{-}& {36.34} & {14.81} & {29.41} & {15.85} \\
		\midrule[0.8pt]
		{\bf T5-CLAPS w/o negative} & {Pos.}& {37.49} & {15.31} & {30.42} & {16.36}  \\
		{\bf T5-CLAPS w/o positive} &{Neg.}& {37.72} & {15.49} & \textbf{30.74} & {16.06}  \\
		{\bf T5-CLAPS} &{Pos.+Neg.}& \textbf{37.89} & \textbf{15.78} & {30.59} & \textbf{16.38}\\
		\midrule[0.8pt]
		{PEGASUS \citep{pegasus}} & {-} & \textbf{47.21} & \textbf{24.56} & {\textbf{39.25}} & {-} \\
		\midrule[0.8pt]
	\end{tabular}
	\vspace{-0.25in}
	\label{summarization}
\end{table}

\subsection{Experimental Results}
\vspace{-0.05in}
\noindent
\textbf{Quantitative Results}
We compare our model with the baseline models on WMT'16 RO-En, XSum, SQuAD dataset for machine translation, text summarization and question generation, respectively. Table \ref{mt-qg} shows that our proposed method CLAPS significantly outperforms the other baseline, with the performance gain of more than $1\%$ on all tasks according to the BLEU scores. Moreover our proposed method improves the performance of the randomly initialized T5 model (Scratch-CLAPS). For question generation, our proposed method also improves F1/EM as well as BLEU scores. It shows that our proposed model is able to generate semantically valid questions that are beneficial for training the QA model.  Note that naively constructing the negative examples for contrastive learning on the both tasks, by randomly shuffling the association of $(\rvx, \rvy)$ from a given mini-batch, degrades the performance. Increasing the batch size to a large value, using larger memory, may increase its performance as observed in SimCLR~\citep{simclr}. However, such an approach will be highly sample-inefficient. On the contrary, our model outperforms all the other baseline models on Xsum dataset for text summarization, as shown in Table \ref{summarization}. For summarization, we observe that contrastive learning with imposters alone can improve the performance by a large margin.

\textbf{Visualization}
To examine our model with proposed contrastive learning framework learns meaningful representation of sentences, we encode a pair of sequences $(\rvx, \rvy)$ into $\mathbf{M}, \mathbf{H}$ with encoder $f$ and decoder $g$. Then, we add perturbations to $\mathbf{H}$ to construct an imposter $\Tilde{\mathbf{H}}$ and an additional positive example $\hat{\mathbf{H}}$ as shown in Eq.~\ref{neg-pert} and \ref{pos-pert1}, \ref{pos-pert2}. We apply average pooling to $\mathbf{M, H,\Tilde{H}, \text{ and } \hat{H}}$ 
and project them onto two dimensional space with t-SNE~\citep{tsne}. As shown in Fig.~\ref{tsne-vis}-(b), the model pushes away the imposter from the embedding of target sequence and pulls the embedding of the distant-targets to the embedding of the source sequence. For the model without contrastive learning, however, the embeddings of both target sequences and distant targets are far away from those of source sequences and the imposters are very close to them as shown in Fig.~\ref{tsne-vis}-(a).

\textbf{Qualitative Examples} For qualitative analysis, we examine the texts that are represented by the distant-target and imposter from our method, CLAPS. To decode them into output sequences, we apply affine transformation and softmax to $\Tilde{\mathbf{H}}$ and $\hat{\mathbf{H}}$ and select the most likely token at each time step. As shown in Table \ref{decoding}, the distant-target example (\textbf{Dist.}), preserves the semantic of the original target sequence (\textbf{GT}) with a single word replaced by a synonym (colored in green). However, the imposters (\textbf{Imp.}) have completely different semantics, and often are gramatically incorrect (colored in red). This shows that the model are exposed to those various valid or incorrect sentences with our proposed contrastive learning framework with adversarial perturbations.

\textbf{Human Evaluation} We further conduct a human evaluation of the 20 summaries and 20 questions generated by our CLAPS and T5-MLE trained for text summarization and QG task.  Specifically, 20 human judges perform blind quality
assessment of two sentences generated by the two models, that are presented in a random order. For text summarization, \textbf{70\%} of the human annotators chose the sentences generated by our model as better than the baseline, and for QG, \textbf{85\%} favored the sentences generated by our model over that of the baseline.

\input{fig-ex}

%% file: fig-ex.tex
\begin{figure}
    \vspace{-0.25in}
    \begin{minipage}{0.50\linewidth}
        \centering
        \vskip 0.2in
        \centerline{\includegraphics[width=1.0\linewidth]{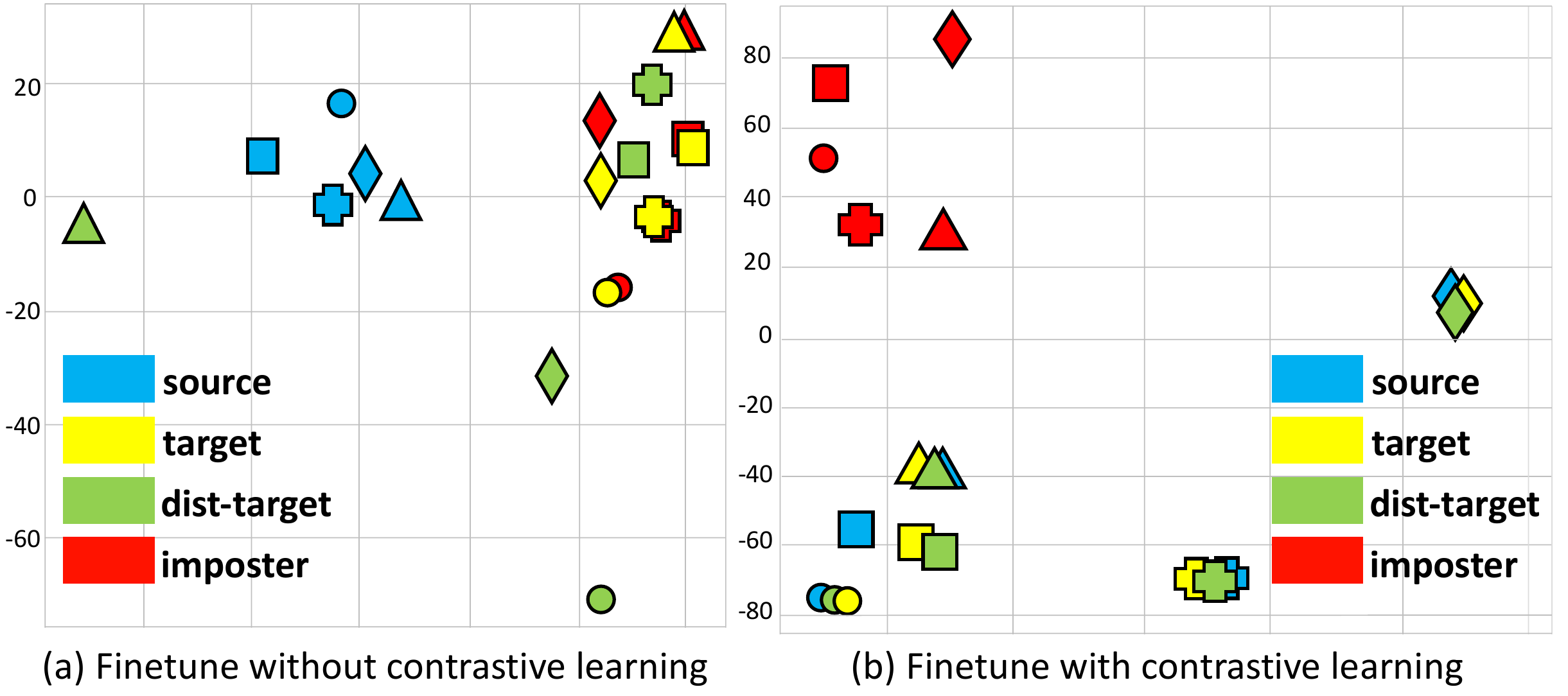}}
        \captionof{figure}{\small \textbf{Visualization.} (a) Embedding space without contrastive learning. (b) Embedding space with our proposed contrastive learning, CLAPS.}
        \label{tsne-vis}
    \end{minipage}
    \hfill
    \begin{minipage}{0.49\linewidth}
        \centering
        \vskip 0.2in
        \centerline{\includegraphics[width=1.0\linewidth]{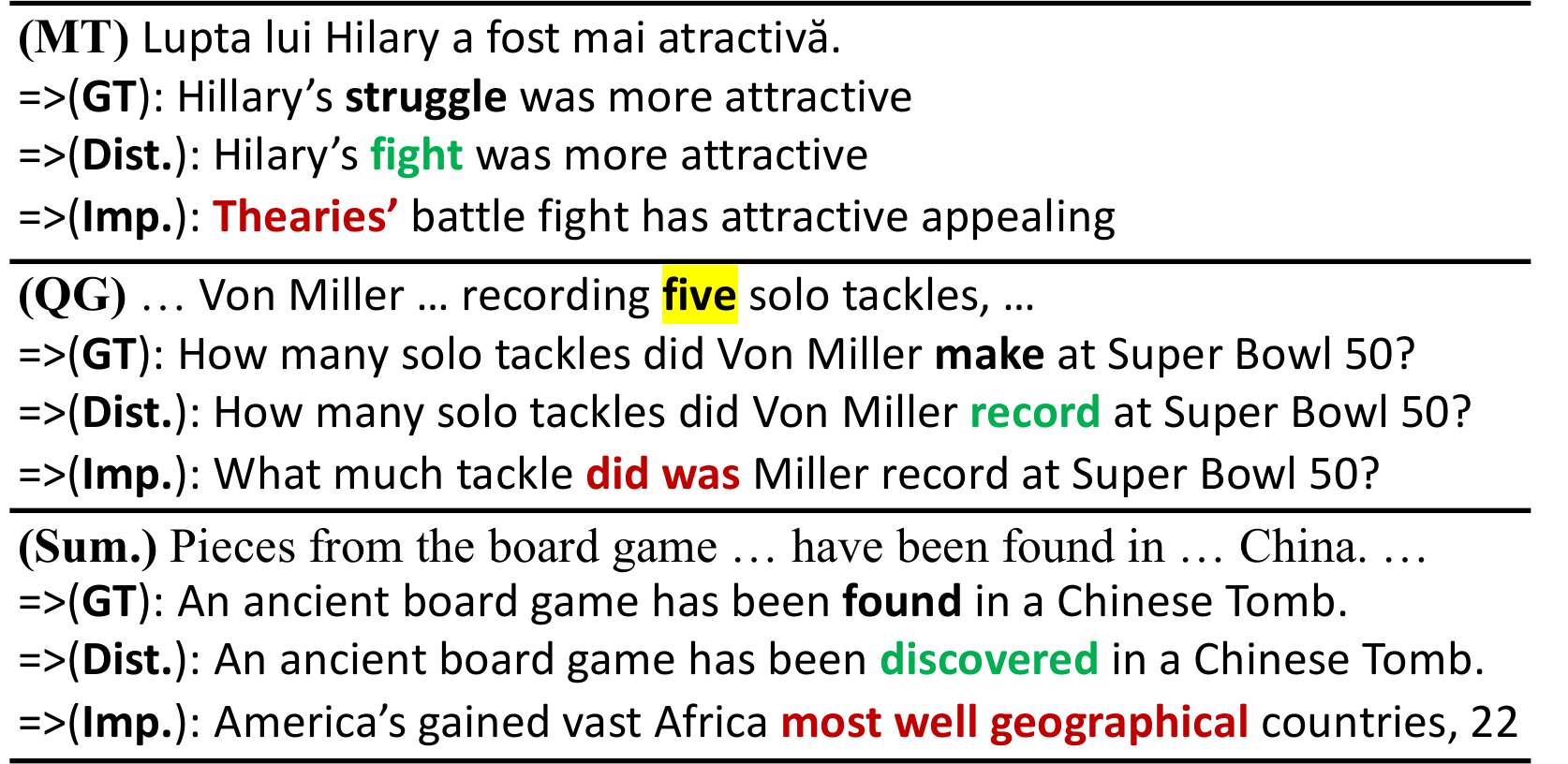}}
        \vspace{-0.15in}
        \captionof{table}{\small Greedy decoding from hidden representation of imposters and distant-targets. The answer span is highlighted for QG.}
        \label{decoding}
    \end{minipage}
    \vskip -0.25in
\end{figure}

%% file: 5conclusion.tex
\section{conclusion}
To mitigate the exposure bias problem in sequence-to-sequence learning, we proposed a contrastive learning framework which maximizes the similarity between ground truth input and output sequence, and minimize the similarity between the input and an incorrect output sequence. Moreover, since conventional approach to sample random non-target examples from the batch as negative examples for contrastive learning results in trivial pairs that are well-discriminated from the beginning, we propose a new principled approach to automatically construct ``hard'' negative and positive examples, where the former is semantically dissimilar but close to the input embedding, and the latter is far from the input embedding but semantically similar. This adversarial learning enables the model to learn both the correct and incorrect variations of the input, and generalize better to unseen inputs. We empirically showed that our method improved the performance of seq2seq model on machine translation, question generation, and text summarization tasks. While we specifically targeted the exposure bias problem with seq2seq models for conditional text generation, our method may be applicable to seq2seq learning for tasks from other domains, such as automatic speech recognition, text-to-speech generation, or video captioning.

\paragraph{Acknowledgements}
This work was supported by Institute of Information \& communications Technology Planning \& Evaluation (IITP) grant funded by the Korea government (MSIT) (No.2020-0-00153), Samsung Advanced Institute of Technology (SAIT), Samsung Electronics Co., Ltd, Institute of Information \& communications Technology Planning \& Evaluation (IITP) grant funded by the Korea government (MSIT)  (No.2019-0-00075, Artificial Intelligence Graduate School Program (KAIST)), and the Engineering Research Center Program through the National Research Foundation of Korea (NRF) funded by the Korean Government MSIT (NRF-2018R1A5A1059921).

%% file: 6appendix.tex
\appendix

\newpage

\begin{table}
	\small
	\centering
	\caption{The statistics and the data source of WMT'16 RO-EN, Xsum, and SQuAD.}
	\begin{tabular}{lllll}
		\toprule[0.8pt]
		\textbf{Datasets} & \textbf{Train (\#)} & \textbf{Valid (\#)} & \textbf{Test (\#)} & \textbf{Source}  \\
		\midrule[0.8pt]
		{WMT'16 RO-EN} & {610,320} & {1,999} & {1,999} & {Romanian-English Parallel corpus.} \\
		\midrule[0.8pt]
		{Xsum} & {204,045} & {11,332} & {11,334} & {One-sentence summary of BBC news articles.} \\
		\midrule[0.8pt]
		{SQuAD} & {86,588} & {5,192} & {5,378} & {Crowd-sourced questions from Wikipedia paragraph}  \\
		\bottomrule[0.8pt]
	\end{tabular}
	\label{datastats}
\end{table}

\section{Experimental Details}
\paragraph{Dataset} For machine translation, text summarization, and question generation, we use WMT'16 RO-EN, Xsum, SQuAD dataset, for each task. The number of train/validation/test set and its source is shown in Table \ref{datastats}. Note that the number of validation and test set for SQuAD is different from the original dataset. Since the original test set is only accessible via the leader board of SQuAD\footnote{\url{https://rajpurkar.github.io/SQuAD-explorer/}}, we split the original validation set into our new validation and test set, following the conventions of question generation communities.

\paragraph{Preprocessing}
For machine translation, we download the raw text\footnote{\url{https://s3.amazonaws.com/datasets.huggingface.co/translation/wmt_en_ro.tar.gz}}, not the tokenized text, and use the same T5-tokenizer as \cite{t5} to tokenize both Romanian and English sentences. We limit the input and output length to 128 tokens. For text summarization, we also use the T5-tokenizer as before, and limit the input length to 512 tokens and output length to 128 tokens. For question generation, we set the maximum length of question as 64 tokens and input which is concatenation of answer and context as 384 tokens. 

\paragraph{Implementation} We finetune the pretrained T5-small model provided from the transformers library~\citep{huggingface}\footnote{\url{https://github.com/huggingface/transformers}} with Adafactor optimizer. We set the batch size 128 and follow the default setting of Adafactor optimizer to finetune the T5-small models. However, the number of negative examples from the batch is 16 or 32 (total batch size divided by the number of GPUs), because we split the batch into smaller batches and distribute them to each GPU machines. We use 8 GPUs for text summarization, and 4 GPUs for machine translation and question generation. The dimension of hidden state of T5 model, $d$ is 512, so we set the hidden size of $\rvz$ as the same. 

\begin{figure}[t]
\vspace{-0.5in}
	\textcolor{black}{\captionof{table}{Translation of Romanian by CLAPS from WMT'16 RO-EN.}}
	\begin{center}
		\includegraphics[width=1.0\linewidth]{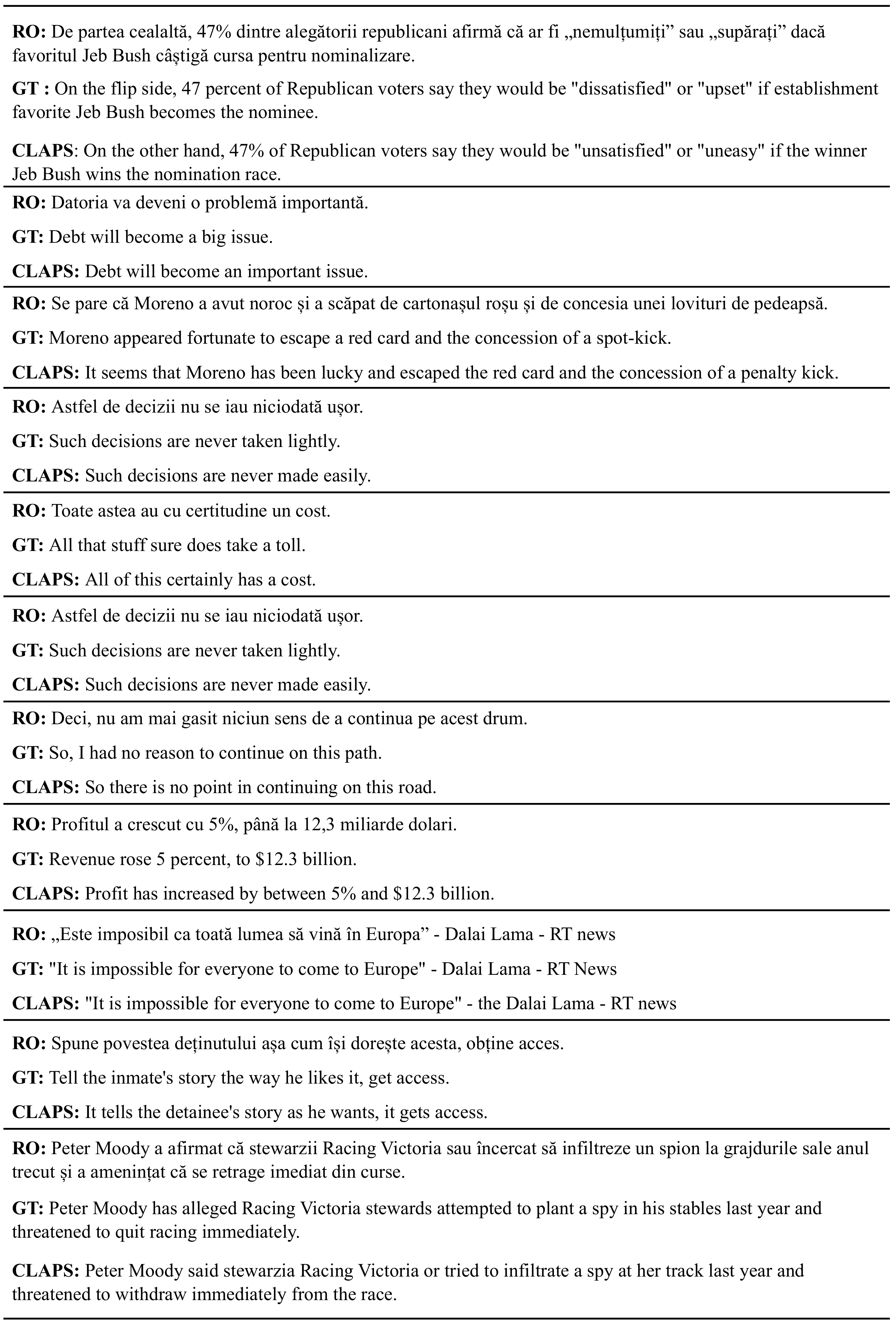}
	\end{center}
	\label{nmt-examples}
	\vspace{-0.20in}
\vspace{-0.12in}
\end{figure}

\paragraph{Evaluation}
We use beam search with beam width 4 to generate the target sentences from the source sentences of the test set. Some of the examples are shown in Table~\ref{sum-examples},\ref{qg-examples},\ref{nmt-examples}. After the generation, we convert the tokens into the raw texts and compare them to the raw text of ground truth target sentences with the automatic evaluation metrics. For n-gram BLEU and Meteor, we use the implementation by \cite{nlgeval}\footnote{\url{https://github.com/Maluuba/nlg-eval}}. For BLEU score, we adopt the implementation by \cite{sacrebleu}\footnote{\url{https://github.com/mjpost/sacrebleu}}. 

\clearpage

\begin{table}[t]
    \centering    
    \caption{Generated summaries by CLAPS from Xsum dataset.}
    \resizebox{0.98\textwidth}{!}{
    \begin{tabular}{l}
    \toprule[1pt]
   \textbf{Article}:  The US military says a strike targeting Taliban in the northern city of \\
   Kunduz may have caused "collateral damage". Offering his "deepest condolences", Mr Obama said \\
   he expected a "full accounting of the facts" and would then make a definitive judgement. …\\
    \\
    \textbf{GT:} President Barack Obama says the US has launched a "full investigation" into air strikes that \\
    killed 19 people at an MSF-run Afghan hospital on Saturday. \\
    \\
    \textbf{CLAPS:} US President Barack Obama has called for an inquiry into air strikes in Afghanistan that\\
    killed dozens of medical workers. \\
    \midrule[1pt]
    \textbf{Article}: Forecasts were for quarterly growth of between 0.5\% and 0.7\%. Official statistics also \\
    showed that household consumption expenditure boosted the quarterly growth numbers. \\But economist Shane Oliver told the BBC the numbers were ``well below potential". \\
    On an annual basis the economy expanded 2.3\%, beating expectations for 2.1\%. \\
    Economic growth in the March quarter of 2014 was 2.9\%. ``The March quarter GDP \\
    $[$gross domestic product$]$ growth was far better than feared just a few days ago," \\
    said Mr Oliver, who is chief economist with AMP Capital in Sydney. \\
    ``However, Australia is still not out of the woods, as annual growth at 2.3\% is well below potential, \\
    and a full 0.8\% percentage points of the 0.9\% growth came from higher inventories and trade." \\
    He said domestic demand remained ``very weak with consumer spending\\
    and home construction only just offsetting the ongoing slump in mining investment". ... \\
    \\
    \textbf{GT}: Australia's economy grew at a better-than-expected 0.9\% in the first quarter of 2015, \\
    compared to the previous quarter, boosted by mining together with financial and insurance services. \\
    \\
    \textbf{CLAPS}: Australia's economy grew faster than expected in the first three months of the year,\\
    according to official figures. \\
    \midrule[1pt]
    \textbf{Article:} After the problems last week, many doubt the system will cope. \\
    Transport for London (TfL) remains confident, although it admits there will be breakdowns. \\ The trick will be in getting the system back up and running quickly. So here's some friendly advice\\ 
    for tourists and Olympic visitors to try and make the transport experience as easy as possible. \\ 
    If anyone thinks of any more please post below. \\
    \\
    \textbf{GT:} The busiest summer ever looms for London's transport system. \\
    \\
    \textbf{CLAPS:} London's transport system has been a pretty busy week. \\
    \midrule[1pt]
    \textbf{Article:} The outgoing vice-president spoke during a state dinner and took the opportunity to praise \\
    America's northern neighbour. "The world is going to spend a lot of time looking to you, \\
    Mr Prime Minister", he told the Canadian leader. Mr Biden has been highly critical of \\
    US President-elect Donald Trump. "Vive le Canada because we need you very, very badly," \\
    he told the dinner guests. He went on to describe the self-doubt that liberal leaders across the world \\
    are currently experiencing after several political defeats. But he praised "genuine leaders" including \\
    German Chancellor Angela Merkel, saying such statesmen and women are in short supply. \\
    Mr Trudeau reportedly became emotional during Mr Biden's remarks when the American \\
    spoke of his late father, former Prime Minister Pierre Trudeau. \\
    "You're a successful father when your children turn out better than you," Mr Biden said. ...\\
    \\
    \textbf{GT:} US Vice-President Joe Biden told an audience in Ottawa that the world needs "genuine leaders" \\
    such as Canadian Prime Minister Justin Trudeau.\\
    \\
    \textbf{CLAPS:} Vice-President Joe Biden has praised Canadian Prime Minister Vive le Canada \\for his visit to the country. \\
    \bottomrule[1pt]
    \end{tabular}
    }
    \label{sum-examples}
\end{table}

\begin{table}[t]
    \centering    
    \resizebox{0.98\textwidth}{!}{
    {\color{black}\begin{tabular}{l}
    \toprule[1pt]
   \textbf{Article}:  The Swedish giant asked customers who bought \\
   any model of the Mysingso chair to return it for a full refund. \\
   The global recall comes after Ikea received reports from \\
   Finland, Germany, the US, Denmark and Australia \\
   that users had received injuries to their fingers that needed medical treatment. \\
   Ikea's statement said the chair had a "risk of falling or finger entrapment". \\
   It said: "After washing the fabric seat it is possible to re-assemble the chair incorrectly\\
   leading to risks of falls or finger entrapments. \\
   "Ikea has received five incident reports in which a Mysingso beach chair collapsed \\
   during use due to incorrect re-assembly. \\
   All five reports included injuries to fingers and required medical attention.\\
   It added that a full investigation had led to an improved design \\
   "to further mitigate the risks of incorrect re-assembly and injuries" \\
   and the updated chair would be available from next month. \\
   Ikea has more than 300 stores in 27 countries.\\
    \\
    \textbf{GT:} Ikea is recalling a beach chair sold in the UK after reports \\
    that it can collapse and cause injury.
    \\
    \\ 
    \textbf{CLAPS:} Ikea is recalling a popular beach chair that collapsed during \\
    use because of incorrect re-assemblies. \\
    \midrule[1pt]
    \textbf{Article:} Spending on the NHS should also be paid for \\
    by a dedicated tax marked on every payslip, the former health minister suggested. \\
    Under Mr Lamb's plan, taxes would not be increased as the new levy would be offset \\
    by deductions to income tax or national insurance. \\
    He has warned the NHS faces collapse without an urgent cash injection. \\
    The plans are not yet party policy and will not be put to \\
    this year's conference in Bournemouth. But Mr Lamb, the party's health spokesman, \\
    told party members he was "very interested in the idea of a dedicated NH \\
    S and care contribution - separating it out from the rest of taxation, \\
    clearly identified on your payslip. "And I am really interested in the idea \\
    of the right for local areas to raise additional funds for the NHS \\
    and care if they choose." The Lib Dems say he would like to implement \\
    the ideas across the UK, although, as health and social care are devolved,\\
    it is unclear how this would be enforced.\\
    Mr Lamb - who lost out to Tim Farron in a leadership election in July\\
    - proposes a cross-party commission to explore the ideas. \\
    He intends to consult health bodies and professionals, \\
    patients, trade unions and academics. Ministers have pledged £2bn in this financial year\\
    for the NHS, and an extra £8bn by 2020. \\
    But Mr Lamb told the BBC that this was insufficient and, \\
    having "seen the books" as a minister in the last government, \\
    he feared the NHS could face a funding shortfall of £30bn by 2020.\\
    "The bottom line is with rising demand because of an ageing population we need more investment," \\
    he said. Mr Lamb also warned that the social care system was "on its knees" \\
    and could collapse without a cash injection of £5bn. \\
    "I've been in the department. I have seen the books and I am deeply concerned. \\
    If we carry on regardless, the system will crash." \\
    Taxpayers are already shown how much they have contributed to the health service\\
    in annual personal tax statements. An attempt to establish a cross-party commission\\
    on social care before the 2010 election - led in part by Mr Lamb - collapsed in acrimony. \\
    \\
    \textbf{GT}: English councils should be allowed to put up taxes to fund the NHS,\\
    Norman Lamb has told the Lib Dem conference. \\
    \\
    \textbf{CLAPS}:A new levy on the NHS and social care should be \\
    introduced by the Liberal Democrats, Norman Lamb has said. \\
    
    \bottomrule[1pt]
    \end{tabular}
    }}
    \label{sum-examples2}
\end{table}

\begin{table}[t]
    \centering    
    \resizebox{0.98\textwidth}{!}{
    {\color{black}\begin{tabular}{l}
    \toprule[1pt]
   \textbf{Article}:  Yorkshire, Lancashire and Derbyshire have been worst affected,\\ 
   after 2-5cm fell overnight, with 10cm reported on higher ground. \\
   Passengers waiting to depart Manchester Airport have reported \\
   being stuck on the runway for hours due to a lack of de-icers. \\
   Leeds Bradford Airport suspended all morning flights but has since reopened.\\
   Manchester Airport reported "minor delays to departing aircraft"\\
   -  but passengers told the BBC they had been stuck on board outbound flights.\\
   Shirley Hale said her Jet2 flight to Tenerife
   \\ had been waiting to depart for over four hours. \\
   "We have been told that there are not enough de-icers at the airport,"\\
   she said. The airport apologised and said de-icing was the responsibility of\\
   airlines and their ground teams. More than 100 schools were closed across East\\
   Lancashire and Oldham, with 80 shut in West Yorkshire. \\
   BBC Weather said Buxton in Derbyshire saw up to 17cm of snow, \\
   the deepest measured on Friday. The avalanche risk in the Peak District was\\
   currently extremely high, Buxton Mountain Rescue Team said.\\
   Parts of Staffordshire have been affected, with several centimetres of \\
   snow reported in Flash, England's highest village. \\
   Commuters have been urged to allow extra journey time, \\
   and the Met Office has issued snow and ice warnings. \\
   More on the snow and other stories in West Yorkshire Weather \\
   updates for Lancashire and Greater Manchester BBC Weather \\
   presenter Kay Crewdson said conditions were due to slowly \\
   improve into Saturday. Molly Greenwood reported 10cm of snow \\
   in the Huddersfield area. "Don't think I'm going anywhere," she said. \\
   Zulfi Hussain said the snow was causing "traffic chaos" in Woodhall Road,\\ Calverley, near Leeds. Elliott Hudson, another West Yorkshire resident,\\
   said: "Looks like I have woken up in Narnia." \\
   West Yorkshire's Liversedge FC, who have had to cancel every home \\
   game for the last four months due to bad weather, \\
   tweeted a picture of snow with the caption: \\
   "It's not looking good for Liversedge FC's home game with Worksop Town tomorrow."\\
   The A628 Woodhead, A57 Snake Pass and A537 Cat and \\
   Fiddle roads are all affected, with delays reported on \\
   the M65 motorway. Highways England said the A57 eastbound \\
   in Great Manchester is closed between M67/A560 and B6174  \\
   due to severe weather conditions. It said teams were working \\
   to clear the road. Tony Hallwood, from Leeds Bradford Airport, \\
   said it reopened at about 09:00 GMT after crews used ploughs \\
   to clear snow from the runway. He said: "We are asking passengers\\
   to make their way to the airport as early as they can given the difficult \\ conditions." Bus operators are also reporting delays to all services\\
   across West Yorkshire. Oldham Council has said 48 schools had closed \\
   this morning as a result of the snow and severe weather. \\
   Drivers are also being asked to take extra care after snow \\
   fell overnight in some parts of Northern Ireland. \\
   A Met Office yellow warning for ice and snow in\\
   northern England and Wales ended at 15:00.
    \\
    \\
    \textbf{GT:} Heavy snowfall has caused travel disruption in parts of northern England.
    \\
    \\ 
    \textbf{CLAPS:} Flights have been disrupted after a large avalanche hit parts of England. \\
    \bottomrule[1pt]
    \end{tabular}}
    }
    \label{sum-examples3}
\end{table}

\begin{table}[t]
    \centering    
    \resizebox{0.98\textwidth}{!}{
    {\color{black}\begin{tabular}{l}
    \toprule[1pt]
   \textbf{Article}:  But once the votes are counted, what can residents expect to pay in council tax?  \\
   Below are the figures for a Band D property for every council area \\
   in Wales for the current financial year of 2017/18, \\
   how much that has gone up by for the current year, \\
   and what the average property in the area actually pays. \\
   They are grouped here by police force region - \\
   council tax includes the police precept which is added to \\
   the overall bill paid by homes. Local government is not fully \\
   funded by council tax. Much of the funding for councils comes \\
   in the form of grants from the Welsh Government, \\
   which in turn gets its funding from the UK government in London. \\
   In 2017/18 a total of £4.1bn is being divided among Wales' 22 councils. \\
   The lions share of council cash goes on schools \\
   - with social services following behind, as shown in the graph above.\\
   Residents pay council tax based on which band their property is in,\\
   based on its worth. Band D has historically been used as \\
   the standard for comparing council tax levels between and across local \\ authorities. It is used to charge tax to a property that, in Wales, \\
   was worth between £91,001 to £123,000 on April 2003 values. \\
   Council tax gets lower the cheaper a property is, \\
   and higher the more expensive a property is. \\
   Council tax figures source: Welsh Government
    \\
    \\
    \textbf{GT:}  Voters will go to the polls on Thursday to determine who will represent them on local councils.
    \\
    \\ 
    \textbf{CLAPS:} The people of Wales are voting in a referendum on whether or not to pay council tax. \\
    \midrule[1pt]
    \textbf{Article: }The side's appearance in France will be its first at a major \\ tournament since the 1958 World Cup.  \\
    Players and coaches left their base at the Vale Resort,\\
    Vale of Glamorgan, on Saturday and headed to Cardiff Airport. \\
    After a send-off from pupils from Ysgol Treganna, Cardiff, \\
    the team took off for a friendly in Sweden on Sunday. \\
    They will then head to France ahead of the team's first \\
    game of the tournament against Slovakia on 11 June. \\
    \\
    \textbf{GT:} Wales' football team has departed the country as their Euro 2016 preparations reach a climax. \\
    \\
    \textbf{CLAPS:} Wales' Euro 2016 squad have arrived in France for the first time since 1958. \\
    \midrule[1pt]
    \textbf{Article: }The 40-year-old, from the South Bank area of Teesside,\\
    was discovered on the A66 in the early hours "in a distressed state"\\
    with wounds to his groin after the attack. \\
    The road, from Greystones Roundabout to Church Lane in Middlesbrough,\\
    was shut earlier while searches of the area were carried out. \\
    It has now reopened. A 22-year-old man was arrested on suspicion of assault\\
    and later bailed. Cleveland Police said the injured man had been \\
    placed in an induced coma in hospital. The force said in a statement: \\
    "Police can confirm that the man found this morning on the A66\\
    had wounds to his groin area. "Officers are continuing to\\
    investigate and are appealing for anyone with information to contact them." \\
    \\
    \textbf{GT: } A man has been found by the side of a road with his penis cut off. \\
    \\
    
    \textbf{CLAPS: } A man is in an induced coma after being found with serious injuries on a Teesside road.\\
    \\
    
    \bottomrule[1pt]
    \end{tabular}}
    }
    \label{sum-examples4}
\end{table}

\begin{table}[t]
    \centering    
    \resizebox{0.98\textwidth}{!}{
    {\color{black}\begin{tabular}{l}
    \toprule[1pt]
   \textbf{Article}:  In July, a major bug was discovered in the software \\
   that could let hijackers access data on up to a billion phones.\\
   Manufacturers have been slow to roll out a fix because many variations \\
   of Android are widely used. One Android expert said it was "about time"\\
   phone makers issued security fixes more quickly. \\
   Android has been working to patch a vulnerability, known as Stagefright,\\
   which could let hackers access a phone's data simply by sending somebody \\
   a video message. "My guess is that this is the single largest software \\
   update the world has ever seen," said Adrian Ludwig, \\
   Android's lead engineer for security, at hacking conference Black Hat.\\
   LG, Samsung and Google have all said a number of their handsets \\
   will get the fix, with further updates every month. \\
   Android is an open source operating system, with the software \\
   freely available for phone manufacturers to modify and use on their handsets.\\
   The Google-led project does provide security fixes for the software, \\
   but phone manufacturers are responsible for sending the updates \\
   to their devices. Some phones running old versions of Android \\
   are no longer updated by the manufacturer. \\
   Many companies also deploy customised versions of Android \\
   which take time to rebuild with the security changes. \\
   Apple and BlackBerry can patch security problems more quickly\\
   because they develop both the software and the hardware for their devices.\\ BlackBerry's software is reviewed by mobile networks before being sent to\\ handsets, while Apple can push updates to its phones whenever it wants. \\
   "The very nature of Android is that manufacturers \\
   add their own software on top, so there have been delays \\
   in software roll-outs," said Jack Parsons, editor of Android Magazine. \\
   "In the US it's even worse because mobile carriers often add their \\
   own software too, adding another layer of bureaucracy holding up security fixes.\\ "There's no real villain here, that's just how the system works. \\
   But there will always be security concerns with software,\\
   so it's right that some of the manufacturers are stepping up to deal with this now."
    \\
    \\
    \textbf{GT:} Samsung, LG and Google have pledged to provide monthly security \\ updates for smartphones running the Android operating system.
    \\
    \\ 
    \textbf{CLAPS:} The world's largest software update is to be issued by Google-led Android. \\
    \midrule[1pt]
    \textbf{Article: }The move follows a claim by Crossmaglen Rangers player\\
    Aaron Cunningham that he was the victim of verbal abuse during \\
    the 2 December Ulster football final. The Ulster Council carried out an \\
    investigation and BBC Sport  understands one Kilcoo player is to \\
    be banned for six months and another for four months. \\
    Kilcoo said they had not been notified, and the players could appeal.\\
    The two suspensions have yet to be officially confirmed \\
    by the Ulster Council. It is believed the case was the \\
    first time an allegation of racial abuse had been lodged\\
    with the provincial governing body. When an investigation was announced,\\
    Ulster GAA president Aogán O Fearghail, said anyone found \\
    guilty of racism would be dealt with severely. \\
    Kilcoo released a statement saying the club condemned \\
    abuse and would co-operate with the Ulster Council's investigation.\\
    The Gaelic Athletic Association, which governs the sport in Ireland,\\
    is to discuss how to deal with racism at its annual congress in March. \\
    \\
    \textbf{GT:} Two Kilcoo players are to be suspended by Ulster GAA chiefs following allegations of racial abuse. \\
    \\
    \textbf{CLAPS:} Two Kilcoo players have been suspended by the Ulster GAA for alleged racial abuse. \\
    \bottomrule[1pt]
    \end{tabular}}
    }
    \label{sum-examples5}
\end{table}

\begin{table}[t]
    \centering
    \caption{Generated Questions by CLAPS from SQuAD. Answer spans are highlighted.}
    \resizebox{0.98\textwidth}{!}{
    \begin{tabular}{l}
    \toprule[1pt]
    \textbf{Context}: ... The Broncos finished the regular season with a 12-4 record, and denied\\
    the New England Patriots a chance to defend their title from Super Bowl XLIX by defeating\\ them 20-18 in the AFC Championship Game. They joined the Patriots, Dallas Cowboys, \\
    and Pittsburgh Steelers as one of four teams that have made \hlyellow{\textbf{eight}} appearances in the Super Bowl.\\
    \\
    \textbf{GT:} How many appearances have the Denver Broncos made in the Super Bowl? \\
    \\
    \textbf{CLAPS:} How many Super Bowl appearances have the Broncos made? \\
    \midrule[1pt]
    \textbf{Context:} In late November 2015, reports surfaced stating that ``multiple acts" would perform \\
    during the halftime show. On December 3, the league confirmed that the show would be \\ headlined by  the \hlyellow{\textbf{British}} rock group Coldplay. On January 7, 2016, Pepsi confirmed to the \\
    Associated Press that Beyoncé, who headlined the Super Bowl XLVII halftime show and\\ collaborated with Coldplay on the single ``Hymn for the Weekend", would be making an appearance. \\
    Bruno Mars, who headlined the Super Bowl XLVIII halftime show, and Mark Ronson also performed.\\
    \\
    \textbf{GT:} What nationality is the band Coldplay? \\
    \\
    \textbf{CLAPS:} What nationality was Coldplay? \\
    \midrule[1pt]
    \textbf{Context:} There are 13 natural reserves in Warsaw - among others, Bielany Forest, Kabaty Woods, \\
    Czerniaków Lake. About 15 kilometres (9 miles) from Warsaw, the Vistula river's environment \\ changes strikingly and features a perfectly preserved ecosystem, with a habitat of animals that \\
    includes the \hlyellow{\textbf{otter, beaver and hundreds of bird species}}. There are also several lakes in Warsaw \\
    - mainly the oxbow lakes, like Czerniaków Lake, the lakes in the Łazienki or Wilanów Parks,\\
    Kamionek Lake. There are lot of small lakes in the parks, but only a few are permanent - the majority \\
    are emptied before winter to clean them of plants and sediments. \\
    \\
    \textbf{GT:} What animals does the Vistula river's ecosystem include? \\
    \\
    \textbf{CLAPS:} What animals are included in the Vistula river's habitat?\\
    \midrule[1pt]
    \textbf{Context:} "The FSO Car Factory was established in 1951. A number of vehicles\\
    have been assembled there over the decades, including the Warszawa, Syrena, Fiat 125p \\
    (under license from Fiat, later renamed 
    FSO 125p when the license expired) and the Polonez. \\
    The last two models listed were also sent abroad and assembled in a number of other countries, \\
    including Egypt and Colombia. In 1995 the factory was purchased by the South Korean \\
    car manufacturer Daewoo, which assembled the Tico, Espero,Nubia, Tacuma, Leganza, Lanos\\
    and Matiz there for the European market. In 2005 the factory was sold to \hlyellow{\textbf{AvtoZAZ}}, \\
    a Ukrainian car manufacturer which assembled there the Chevrolet Aveo. \\
    The license for the production of the Aveo expired in February 2011 and has since not been\\ renewed. Currently the company is defunct." \\
    \\
    \textbf{GT:} Who bought the factory in 2005? \\
    \\
    \textbf{CLAPS:} To whom was the factory sold in 2005? \\
    \midrule[1pt]
    \textbf{Context:} The Scotland Act 1998, which was passed by the Parliament\\
    of the United Kingdom and given royal assent by Queen Elizabeth II on 19 November 1998,\\
    governs the functions and role of the Scottish Parliament and delimits its legislative competence. \\
    The Scotland Act 2012 extends the \hlyellow{\textbf{devolved competencies}.} \\
    For the purposes of parliamentary sovereignty, the Parliament of the United Kingdom \\
    at Westminster continues to constitute the supreme legislature of Scotland. \\
    However, under the terms of the Scotland Act,\\
    Westminster agreed to devolve some of its responsibilities over \\
    Scottish domestic policy to the Scottish Parliament.\\
    Such \"devolved matters\" include education, health, agriculture and justice. \\
    The Scotland Act enabled the Scottish Parliament to pass primary legislation on these issues.\\
    A degree of domestic authority, and all foreign policy, remain with the UK Parliament in Westminster. \\
    The Scottish Parliament has the power to pass laws and has limited tax-varying capability. \\
    Another of the roles of the Parliament is to hold the Scottish Government to account. \\
    \\
    \textbf{GT: } What does the Scotland Act of 2012 extend? \\
    \\
    \text{CLAPS: } What does the Scotland Act 2012 extend? \\
    \bottomrule[1pt]
    \end{tabular}
    }
    \label{qg-examples}
\end{table}

\begin{table}[]
    \centering
    \resizebox{0.98\textwidth}{!}{
    {\color{black}\begin{tabular}{l}
    \toprule[1pt]
    \textbf{Context: }
    Stage 1 is the first, or introductory stage of the bill, \\
    where the minister or member in charge of the bill will formally introduce it to Parliament together \\
    with its accompanying documents-Explanatory Notes, a Policy Memorandum \\
    setting out the policy underlying the bill, \\
    and a Financial Memorandum setting out the costs and savings associated with it.\\
    Statements from the Presiding Officer and\\
    the member in charge of the bill are also lodged indicating whether the bill is \\
    within the legislative competence of the Parliament. \\
    Stage 1 usually takes place, initially, in the relevant committee or committees \\
    and is then submitted to \hlyellow{\textbf{the whole Parliament}} \\
    for a full debate in the chamber on the general principles of the bill. \\
    If the whole Parliament agrees in a vote to the general principles of the bill, it then proceeds to Stage 2. \\
    \\
    \textbf{GT: } Where are bills typically gestated in Stage 1? \\
    \\
    \textbf{CLAPS: }Where does Stage 1 usually take place? \\
    \midrule[1pt]
    
    \textbf{Context: } Moderate and reformist Islamists who accept and \\
    work within the democratic process include parties like the Tunisian Ennahda Movement.\\ Jamaat-e-Islami of Pakistan is basically a socio-political \\
    and democratic Vanguard party but has also gained political influence \\
    through military coup d'état in past. \\
    The Islamist groups like Hezbollah in Lebanon and Hamas in \hlyellow{\textbf{Palestine}} participate \\
    in democratic and political process as well as armed attacks, \\
    seeking to abolish the state of Israel. \\
    Radical Islamist organizations like al-Qaeda and the Egyptian Islamic Jihad, \\
    and groups such as the Taliban, entirely reject democracy,\\
    often declaring as kuffar those Muslims who support it (see takfirism), \\
    as well as calling for violent/offensive jihad \\
    or urging and conducting attacks on a religious basis.\\
    \\
    \textbf{GT: } Where does Hamas originate?\\
    \\
    \textbf{CLAPS: }Where are Hamas located?\\
    \midrule[1pt]
    \textbf{Context: }Sayyid Abul Ala Maududi was an important early twentieth-century figure \\
    in the Islamic revival in India, and then after independence from Britain, in Pakistan. \\
    Trained as a lawyer he chose the profession of journalism, \\
    and wrote about contemporary issues and most importantly about Islam and Islamic law. \\
    Maududi founded the Jamaat-e-Islami party in 1941 and 
    remained its leader until 1972. \\
    However, Maududi had much more impact \hlyellow{\textbf{through his writing}}\\
    than through his political organising. His extremely influential books (translated into many languages) \\
    placed Islam in a modern context, and influenced not only conservative ulema \\
    but liberal modernizer Islamists such as al-Faruqi, \\
    whose \"Islamization of Knowledge\" carried forward some of Maududi's key principles. \\
    \\
    \textbf{GT:} Where did Maududi exert the most impact? \\
    \\
    \textbf{CLAPS:} How did Maududi have more impact on Islam than his political organising? \\
    \midrule[1pt]
    \textbf{Context: } ByLike many other mainline Protestant denominations in the United States, \\
    the United Methodist Church has experienced significant membership losses in recent decades. \\
    At the time of its formation, the UMC had about 11 million members \\
    in nearly \hlyellow{\textbf{42,000}} congregations. \\
    In 1975, membership dropped below 10 million for the first time. \\
    In 2005, there were about 8 million members in over 34,000 congregations. \\
    Membership is concentrated primarily in the Midwest and in the South. \\
    Texas has the largest number of members, with about 1 million. \\
    The states with the highest membership rates are \\
    Oklahoma, Iowa, Mississippi, West Virginia, and North Carolina.
    \\ \\
    \textbf{GT: } At the time of its formation, how many congregations did the UMC have?

    \\
    \textbf{CLAPS: }How many congregations did the UMC have at the time of its formation?

    \\
    \bottomrule[1pt]
    \end{tabular}}
    }
\end{table}

\begin{table}[t]
    \centering
    \resizebox{0.98\textwidth}{!}{
    {\color{black}\begin{tabular}{l}
    \toprule[1pt]
    
    \textbf{Context: }
    Celoron's expedition force consisted of about 200 Troupes de la marine and 30 Indians. \\
    The expedition covered about 3,000 miles (4,800 km) between June and November 1749. \\
    It went up the St. Lawrence, continued along the northern shore of Lake Ontario, \\
    crossed the portage at Niagara, and followed the southern shore of Lake Erie. \\
    At the Chautauqua Portage (near present-day Barcelona, New York), \\
    the expedition moved inland to the Allegheny River, which it followed to the site of present-day Pittsburgh. \\
    There Céloron buried lead plates engraved with the French claim to the Ohio Country. \\
    Whenever he encountered British merchants or fur-traders, \\
    \hlyellow{\textbf{Celoron informed them of the French claims on the territory and told them to leave.}}
    \\
    \\
    \textbf{GT: }How did Celeron handle business on trip?\\
    \\
    \textbf{CLAPS: }What did Celoron do when he encountered the British?\\
    \midrule[1pt]
    \textbf{Context: }Like many cities in Central and Eastern Europe, infrastructure in Warsaw \\
         suffered considerably during its time as an Eastern Bloc economy\\
         - though it is worth mentioning that the initial Three-Year Plan \\
         to rebuild Poland (especially Warsaw) was a major success, \\
         but what followed was very much the opposite. \\
         However, over the past decade Warsaw has seen many improvements \\
         due to solid economic growth, an increase in foreign investment \\
         as well as funding from the European Union. In particular,\\
         the city's metro, roads, sidewalks, health care facilities \\
         and sanitation facilities have \hlyellow{\textbf{improved markedly}}.
answer:improved markedly\\
\\
    
    \textbf{GT: } Warsaw's sidewalks and sanitation facilities are some examples of things which have what?\\
    \\
    \textbf{CLAPS: } What has happened to Warsaw's infrastructure in the past decade? \\
    \midrule[1pt]
    \textbf{Context: }Several commemorative events take place every year. \\
    Gatherings of \hlyellow{\textbf{thousands}} of people on the banks of the Vistula \\
    on Midsummer's Night for a festival called Wianki (Polish for Wreaths)\\
    have become a tradition and a yearly event in the programme of cultural events\\
    in Warsaw. The festival traces its roots to a peaceful pagan ritual \\
    where maidens would float their wreaths of herbs on the water to predict\\
    when they would be married, and to whom. \\
    By the 19th century this tradition had become a festive event, \\
    and it continues today. The city council organize concerts and other events.\\
    Each Midsummer's Eve, apart from the official floating of wreaths,\\
    jumping over fires, looking for the fern flower, \\
    there are musical performances, dignitaries' speeches, fairs\\
    and fireworks by the river bank. \\
    \\
    \textbf{GT: } How man people gather along the banks of the Vistula for the Wianki festival?\\
    \\
    \textbf{CLAPS: }How many people gather on the banks of the Vistula on Midsummer's Night\\
    for a festival called Wianki?\\
    \midrule[1pt]
    \textbf{Context: }The origin of the legendary figure is not fully known. \\
    The best-known legend, by Artur Oppman, is that long ago two of Triton's \\ daughters set out on a journey through the depths of the oceans and seas.\\
    One of them decided to stay on the coast of Denmark and can be seen sitting\\
    at the entrance to the port of Copenhagen. \\
    The second mermaid reached the mouth of the Vistula River and\\
    plunged into its waters. She stopped to rest on a sandy beach \\
    by the village of Warszowa, where fishermen came to admire her beauty\\
    and listen to her beautiful voice. A greedy merchant also heard her songs;\\
    he followed the fishermen and \hlyellow{\textbf{captured}} the mermaid.\\
    \\
    \textbf{GT: } What did a greedy merchant do to the mermaid? \\
    \\
    \textbf{CLAPS: } What did Oppman do to the mermaid? \\
    \bottomrule[1pt]
    \end{tabular}}
    }
\end{table}

\begin{table}[]
    \centering
    \resizebox{0.98\textwidth}{!}{
    {\color{black}\begin{tabular}{l}
    \toprule[1pt]
     \textbf{Context: }Warsaw remained the capital of the Polish-Lithuanian\\ Commonwealth \hlyellow{\textbf{until 1796}}, when it was annexed by the Kingdom of Prussia\\
    to become the capital of the province of South Prussia. \\
    Liberated by Napoleon's army in 1806, Warsaw was made \\
    the capital of the newly created Duchy of Warsaw. \\
    Following the Congress of Vienna of 1815, Warsaw became the centre of \\
    the Congress Poland, a constitutional monarchy under a personal union \\
    with Imperial Russia. The Royal University of Warsaw was established in 1816.\\
    \\
    \textbf{GT: }How long was Warsaw the capital of the Polish-Lithuanian Commonwealth? \\
    \\
    \textbf{CLAPS: } How long did Warsaw remain the capital of the Polish-Lithuanian Commonwealth? \\
    \midrule[1pt]
    \textbf{Context:} John Paul II's visits to his native country in 1979\\
    and 1983 brought support to the budding solidarity movement \\
    and encouraged the \hlyellow{\textbf{growing anti-communist}} fervor there. \\
    In 1979, less than a year after becoming pope, John Paul celebrated \\
    Mass in Victory Square in Warsaw and ended his sermon with a call to \\
    "renew the face" of Poland: Let Thy Spirit descend! \\
    Let Thy Spirit descend and renew the face of the land! \\
    This land! These words were very meaningful for the Polish citizens\\
    who understood them as the incentive for the democratic changes. \\
    \textbf{GT: } What is St. John's Cathedral an example of, stylistically?\\
    \\
    \textbf{CLAPS: } St. John's Cathedral is a typical example of what style?\\
    \midrule[1pt]
    \textbf{Context: }Gothic architecture is represented in the majestic churches\\
    but also at the burgher houses and fortifications. \\
    The most significant buildings are St. John's Cathedral (14th century), \\
    the temple is a typical example of the so-called \hlyellow{\textbf{Masovian gothic style}}, \\
    St. Mary's Church (1411), a town house of Burbach family (14th century),\\
    Gunpowder Tower (after 1379) and the Royal Castle Curia Maior (1407\u20131410). \\
    The most notable examples of Renaissance architecture in the city \\
    are the house of Baryczko merchant family (1562), building called "The Negro"\\
    (early 17th century) and Salwator tenement (1632). The most interesting examples \\
    of mannerist architecture are the Royal Castle (1596\u20131619) and \\
    the Jesuit Church (1609\u20131626) at Old Town. \\
    Among the first structures of the early baroque the most important \\
    are St. Hyacinth's Church (1603-1639) and Sigismund's Column (1644). \\
    \\
    \textbf{GT: } What is St. John's Cathedral an example of, stylistically?\\
    \\
    \textbf{CLAPS: } St. John's Cathedral is a typical example of what style?
    \\
    \bottomrule[1pt]
    \end{tabular}}
    }
\end{table}

\begin{figure}[t]
\vspace{-0.3in}
	\begin{center}
		\includegraphics[width=1.0\linewidth]{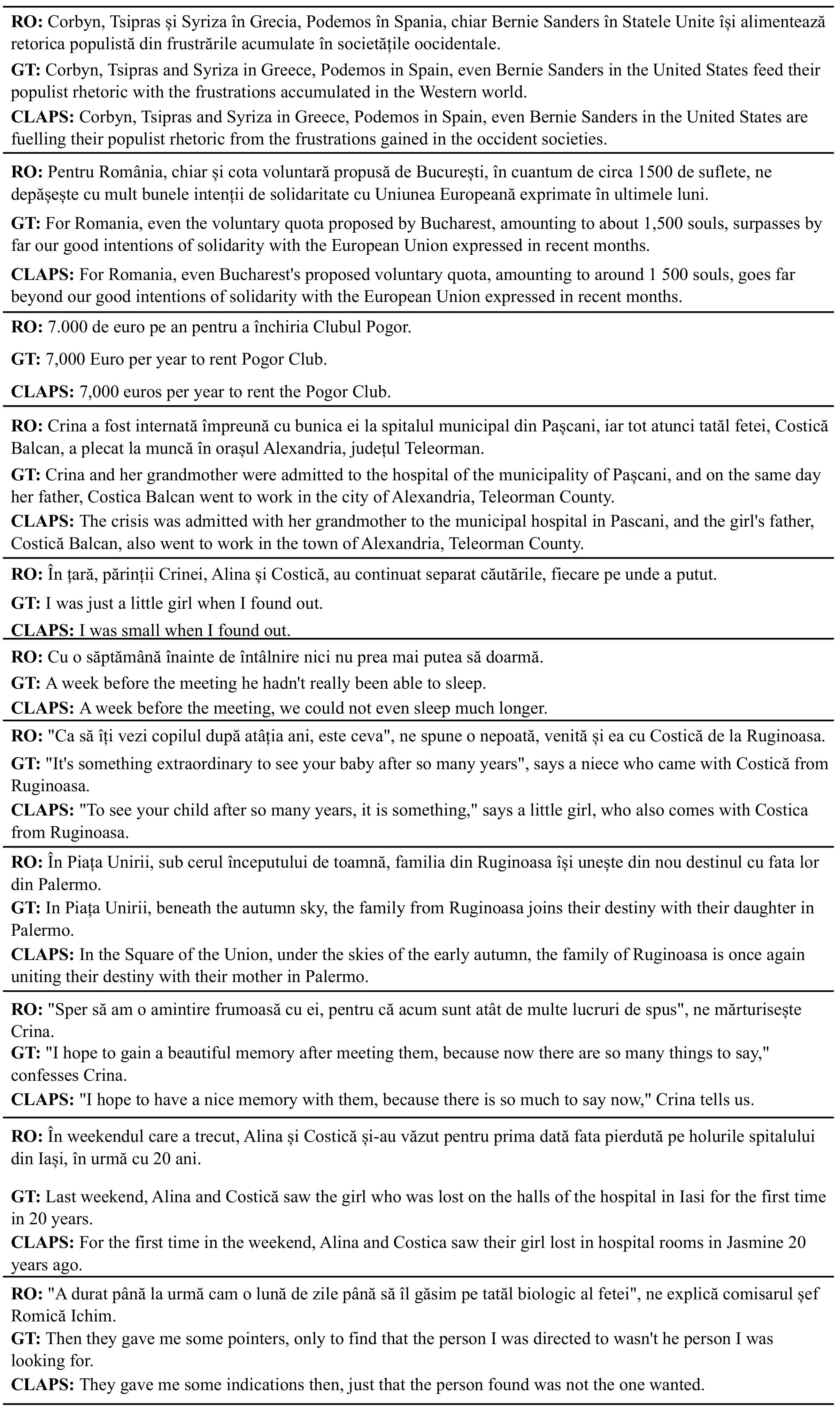}
	\end{center}
	\label{nmt-examples2}
	\vspace{-0.20in}
\vspace{-0.12in}
\end{figure}

%% file: main.bbl
\begin{thebibliography}{52}
\providecommand{\natexlab}[1]{#1}
\providecommand{\url}[1]{\texttt{#1}}
\expandafter\ifx\csname urlstyle\endcsname\relax
  \providecommand{\doi}[1]{doi: #1}\else
  \providecommand{\doi}{doi: \begingroup \urlstyle{rm}\Url}\fi

\bibitem[Aghajanyan et~al.(2020)Aghajanyan, Shrivastava, Gupta, Goyal,
  Zettlemoyer, and Gupta]{better-finetuning}
Armen Aghajanyan, Akshat Shrivastava, Anchit Gupta, Naman Goyal, Luke
  Zettlemoyer, and Sonal Gupta.
\newblock Better fine-tuning by reducing representational collapse.
\newblock \emph{arXiv preprint arXiv:2008.03156}, 2020.

\bibitem[Bahdanau et~al.(2015)Bahdanau, Cho, and Bengio]{attention}
Dzmitry Bahdanau, Kyunghyun Cho, and Yoshua Bengio.
\newblock Neural machine translation by jointly learning to align and
  translate.
\newblock \emph{International Conference on Learning Representations, {ICLR}},
  2015.

\bibitem[Bahdanau et~al.(2017)Bahdanau, Brakel, Xu, Goyal, Lowe, Pineau,
  Courville, and Bengio]{rl-mt}
Dzmitry Bahdanau, Philemon Brakel, Kelvin Xu, Anirudh Goyal, Ryan Lowe, Joelle
  Pineau, Aaron~C. Courville, and Yoshua Bengio.
\newblock An actor-critic algorithm for sequence prediction.
\newblock \emph{International Conference on Learning Representations, {ICLR}},
  2017.

\bibitem[Banerjee \& Lavie(2005)Banerjee and Lavie]{meteor}
Satanjeev Banerjee and Alon Lavie.
\newblock Meteor: An automatic metric for mt evaluation with improved
  correlation with human judgments.
\newblock \emph{Proceedings of the acl workshop on intrinsic and extrinsic
  evaluation measures for machine translation and/or summarization}, 2005.

\bibitem[Bengio et~al.(2015)Bengio, Vinyals, Jaitly, and
  Shazeer]{scheduled-sampling}
Samy Bengio, Oriol Vinyals, Navdeep Jaitly, and Noam Shazeer.
\newblock Scheduled sampling for sequence prediction with recurrent neural
  networks.
\newblock In \emph{Advances in Neural Information Processing Systems}, pp.\
  1171--1179, 2015.

\bibitem[Caccia et~al.(2019)Caccia, Caccia, Fedus, Larochelle, Pineau, and
  Charlin]{gan-falling-short}
Massimo Caccia, Lucas Caccia, William Fedus, Hugo Larochelle, Joelle Pineau,
  and Laurent Charlin.
\newblock Language gans falling short.
\newblock In \emph{International Conference on Learning Representations}, 2019.

\bibitem[Chen et~al.(2020)Chen, Kornblith, Norouzi, and Hinton]{simclr}
Ting Chen, Simon Kornblith, Mohammad Norouzi, and Geoffrey Hinton.
\newblock A simple framework for contrastive learning of visual
  representations.
\newblock \emph{International Conference on Machine Learning, {ICML}}, 2020.

\bibitem[Chopra et~al.(2005)Chopra, Hadsell, and LeCun]{triplet1}
Sumit Chopra, Raia Hadsell, and Yann LeCun.
\newblock Learning a similarity metric discriminatively, with application to
  face verification.
\newblock \emph{2005 IEEE Computer Society Conference on Computer Vision and
  Pattern Recognition (CVPR'05)}, 2005.

\bibitem[Choshen et~al.(2020)Choshen, Fox, Aizenbud, and Abend]{rl-not-good}
Leshem Choshen, Lior Fox, Zohar Aizenbud, and Omri Abend.
\newblock On the weaknesses of reinforcement learning for neural machine
  translation.
\newblock \emph{International Conference on Learning Representations, {ICLR}},
  2020.

\bibitem[Conneau \& Lample(2019)Conneau and Lample]{clm}
Alexis Conneau and Guillaume Lample.
\newblock Cross-lingual language model pretraining.
\newblock In \emph{Advances in Neural Information Processing Systems}, pp.\
  7059--7069, 2019.

\bibitem[Dong et~al.(2019)Dong, Yang, Wang, Wei, Liu, Wang, Gao, Zhou, and
  Hon]{unilm}
Li~Dong, Nan Yang, Wenhui Wang, Furu Wei, Xiaodong Liu, Yu~Wang, Jianfeng Gao,
  Ming Zhou, and Hsiao-Wuen Hon.
\newblock Unified language model pre-training for natural language
  understanding and generation.
\newblock \emph{Advances in Neural Information Processing Systems}, 2019.

\bibitem[Du \& Cardie(2018)Du and Cardie]{harvesting}
Xinya Du and Claire Cardie.
\newblock Harvesting paragraph-level question-answer pairs from wikipedia.
\newblock \emph{Annual Meeting of the Association for Computational
  Linguistics, {ACL}}, 2018.

\bibitem[Goodfellow et~al.(2015)Goodfellow, Shlens, and Szegedy]{adv-example}
Ian~J. Goodfellow, Jonathon Shlens, and Christian Szegedy.
\newblock Explaining and harnessing adversarial examples.
\newblock \emph{International Conference on Learning Representations, {ICLR}},
  2015.

\bibitem[Gutmann \& Hyv{\"a}rinen(2012)Gutmann and Hyv{\"a}rinen]{nce}
Michael~U Gutmann and Aapo Hyv{\"a}rinen.
\newblock Noise-contrastive estimation of unnormalized statistical models, with
  applications to natural image statistics.
\newblock \emph{The journal of machine learning research}, 2012.

\bibitem[Huang et~al.(2018)Huang, Li, Ping, and Huang]{margin-lm}
Jiaji Huang, Yi~Li, Wei Ping, and Liang Huang.
\newblock Large margin neural language model.
\newblock \emph{Empirical Methods in Natural Language Processing, {EMNLP}},
  2018.

\bibitem[Jang et~al.(2017)Jang, Gu, and Poole]{gumbelsoftmax1}
Eric Jang, Shixiang Gu, and Ben Poole.
\newblock Categorical reparameterization with gumbel-softmax.
\newblock \emph{International Conference on Learning Representations, {ICLR}},
  2017.

\bibitem[Jiang et~al.(2020)Jiang, He, Chen, Liu, Gao, and Zhao]{smart}
Haoming Jiang, Pengcheng He, Weizhu Chen, Xiaodong Liu, Jianfeng Gao, and Tuo
  Zhao.
\newblock Smart: Robust and efficient fine-tuning for pre-trained natural
  language models through principled regularized optimization.
\newblock \emph{ACL}, 2020.

\bibitem[Lee et~al.(2020)Lee, Lee, Jeong, Kim, and Hwang]{info-hcvae}
Dong~Bok Lee, Seanie Lee, Woo~Tae Jeong, Donghwan Kim, and Sung~Ju Hwang.
\newblock Generating diverse and consistent {QA} pairs from contexts with
  information-maximizing hierarchical conditional vaes.
\newblock In \emph{Proceedings of the 58th Annual Meeting of the Association
  for Computational Linguistics, {ACL} 2020, Online, July 5-10, 2020}, pp.\
  208--224. Association for Computational Linguistics, 2020.

\bibitem[Lewis et~al.(2020)Lewis, Liu, Goyal, Ghazvininejad, Mohamed, Levy,
  Stoyanov, and Zettlemoyer]{bart}
Mike Lewis, Yinhan Liu, Naman Goyal, Marjan Ghazvininejad, Abdelrahman Mohamed,
  Omer Levy, Veselin Stoyanov, and Luke Zettlemoyer.
\newblock {BART:} denoising sequence-to-sequence pre-training for natural
  language generation, translation, and comprehension.
\newblock \emph{Annual Meeting of the Association for Computational
  Linguistics, {ACL}}, 2020.

\bibitem[Lin \& Hovy(2002)Lin and Hovy]{rouge}
Chin-Yew Lin and Eduard Hovy.
\newblock Manual and automatic evaluation of summaries.
\newblock \emph{ACL Workshop on Automatic Summarization}, 2002.

\bibitem[Liu \& Sun(2015)Liu and Sun]{alignment}
Yang Liu and Maosong Sun.
\newblock Contrastive unsupervised word alignment with non-local features.
\newblock \emph{Proceedings of the Twenty-Ninth {AAAI} Conference on Artificial
  Intelligence,}, 2015.

\bibitem[Logeswaran \& Lee(2018)Logeswaran and Lee]{quick-thought}
Lajanugen Logeswaran and Honglak Lee.
\newblock An efficient framework for learning sentence representations.
\newblock \emph{International Conference on Learning Representations}, 2018.

\bibitem[Maaten \& Hinton(2008)Maaten and Hinton]{tsne}
Laurens van~der Maaten and Geoffrey Hinton.
\newblock Visualizing data using t-sne.
\newblock \emph{Journal of machine learning research}, 2008.

\bibitem[Madry et~al.(2018)Madry, Makelov, Schmidt, Tsipras, and Vladu]{pgd}
Aleksander Madry, Aleksandar Makelov, Ludwig Schmidt, Dimitris Tsipras, and
  Adrian Vladu.
\newblock Towards deep learning models resistant to adversarial attacks.
\newblock In \emph{International Conference on Learning Representations}, 2018.

\bibitem[Mao et~al.(2016)Mao, Huang, Toshev, Camburu, Yuille, and
  Murphy]{cap-gen1}
Junhua Mao, Jonathan Huang, Alexander Toshev, Oana Camburu, Alan~L. Yuille, and
  Kevin Murphy.
\newblock Generation and comprehension of unambiguous object descriptions.
\newblock \emph{2016 {IEEE} Conference on Computer Vision and Pattern
  Recognition, {CVPR}}, 2016.

\bibitem[Mikolov et~al.(2013)Mikolov, Sutskever, Chen, Corrado, and Dean]{w2v}
Tomas Mikolov, Ilya Sutskever, Kai Chen, Greg~S Corrado, and Jeff Dean.
\newblock Distributed representations of words and phrases and their
  compositionality.
\newblock \emph{Advances in neural information processing systems}, 2013.

\bibitem[Miyato et~al.(2017)Miyato, Dai, and Goodfellow]{virtual}
Takeru Miyato, Andrew~M. Dai, and Ian~J. Goodfellow.
\newblock Adversarial training methods for semi-supervised text classification.
\newblock \emph{International Conference on Learning Representations, {ICLR}},
  2017.

\bibitem[Nair \& Hinton(2010)Nair and Hinton]{relu}
Vinod Nair and Geoffrey~E Hinton.
\newblock Rectified linear units improve restricted boltzmann machines.
\newblock \emph{ICML}, 2010.

\bibitem[Narayan et~al.(2018)Narayan, Cohen, and Lapata]{xsum}
Shashi Narayan, Shay~B Cohen, and Mirella Lapata.
\newblock Don’t give me the details, just the summary! topic-aware
  convolutional neural networks for extreme summarization.
\newblock \emph{Empirical Methods in Natural Language Processing, {EMNLP}},
  2018.

\bibitem[Ng et~al.(2020)Ng, Cho, and Ghassemi]{ssmba}
Nathan Ng, Kyunghyun Cho, and Marzyeh Ghassemi.
\newblock Ssmba: Self-supervised manifold based data augmentation for improving
  out-of-domain robustness.
\newblock \emph{Empirical Methods in Natural Language Processing, {EMNLP}},
  2020.

\bibitem[Papineni et~al.(2002)Papineni, Roukos, Ward, and Zhu]{bleu}
Kishore Papineni, Salim Roukos, Todd Ward, and Wei{-}Jing Zhu.
\newblock Bleu: a method for automatic evaluation of machine translation.
\newblock \emph{Proceedings of the 40th Annual Meeting of the Association for
  Computational Linguistics, {ACL} 2002}, 2002.

\bibitem[Paulus et~al.(2018)Paulus, Xiong, and Socher]{rl-sum}
Romain Paulus, Caiming Xiong, and Richard Socher.
\newblock A deep reinforced model for abstractive summarization.
\newblock \emph{International Conference on Learning Representations, {ICLR}},
  2018.

\bibitem[Post(2018)]{sacrebleu}
Matt Post.
\newblock A call for clarity in reporting bleu scores.
\newblock \emph{Proceedings of the Third Conference on Machine Translation:
  Research Papers}, 2018.

\bibitem[Raffel et~al.(2020)Raffel, Shazeer, Roberts, Lee, Narang, Matena,
  Zhou, Li, and Liu]{t5}
Colin Raffel, Noam Shazeer, Adam Roberts, Katherine Lee, Sharan Narang, Michael
  Matena, Yanqi Zhou, Wei Li, and Peter~J. Liu.
\newblock Exploring the limits of transfer learning with a unified text-to-text
  transformer.
\newblock \emph{Journal of Machine Learning Research}, 2020.

\bibitem[Rajpurkar et~al.(2016)Rajpurkar, Zhang, Lopyrev, and Liang]{squad}
Pranav Rajpurkar, Jian Zhang, Konstantin Lopyrev, and Percy Liang.
\newblock Squad: 100,000+ questions for machine comprehension of text.
\newblock \emph{Empirical Methods in Natural Language Processing, {EMNLP}},
  2016.

\bibitem[Ranzato et~al.(2016)Ranzato, Chopra, Auli, and Zaremba]{exposure-bias}
Marc'Aurelio Ranzato, Sumit Chopra, Michael Auli, and Wojciech Zaremba.
\newblock Sequence level training with recurrent neural networks.
\newblock \emph{nternational Conference on Learning Representations, {ICLR}},
  2016.

\bibitem[Schroff et~al.(2015)Schroff, Kalenichenko, and Philbin]{triplet3}
Florian Schroff, Dmitry Kalenichenko, and James Philbin.
\newblock Facenet: A unified embedding for face recognition and clustering.
\newblock \emph{Proceedings of the IEEE conference on computer vision and
  pattern recognition}, 2015.

\bibitem[See et~al.(2017)See, Liu, and Manning]{pointer-generator}
Abigail See, Peter~J Liu, and Christopher~D Manning.
\newblock Get to the point: Summarization with pointer-generator networks.
\newblock \emph{Annual Meeting of the Association for Computational
  Linguistics, {ACL}}, 2017.

\bibitem[Sharma et~al.(2017)Sharma, El~Asri, Schulz, and Zumer]{nlgeval}
Shikhar Sharma, Layla El~Asri, Hannes Schulz, and Jeremie Zumer.
\newblock Relevance of unsupervised metrics in task-oriented dialogue for
  evaluating natural language generation.
\newblock \emph{CoRR}, 2017.

\bibitem[Shazeer \& Stern(2018)Shazeer and Stern]{adafactor}
Noam Shazeer and Mitchell Stern.
\newblock Adafactor: Adaptive learning rates with sublinear memory cost.
\newblock \emph{arXiv preprint arXiv:1804.04235}, 2018.

\bibitem[Sutskever et~al.(2014)Sutskever, Vinyals, and Le]{seq2seq}
Ilya Sutskever, Oriol Vinyals, and Quoc~V Le.
\newblock Sequence to sequence learning with neural networks.
\newblock \emph{Advances in neural information processing systems}, 2014.

\bibitem[Vaswani et~al.(2017)Vaswani, Shazeer, Parmar, Uszkoreit, Jones, Gomez,
  Kaiser, and Polosukhin]{transformer}
Ashish Vaswani, Noam Shazeer, Niki Parmar, Jakob Uszkoreit, Llion Jones,
  Aidan~N Gomez, {\L}ukasz Kaiser, and Illia Polosukhin.
\newblock Attention is all you need.
\newblock \emph{Advances in neural information processing systems}, 2017.

\bibitem[Vedantam et~al.(2017)Vedantam, Bengio, Murphy, Parikh, and
  Chechik]{cap-gen2}
Ramakrishna Vedantam, Samy Bengio, Kevin Murphy, Devi Parikh, and Gal Chechik.
\newblock Context-aware captions from context-agnostic supervision.
\newblock \emph{{IEEE} Conference on Computer Vision and Pattern Recognition,
  {CVPR} 2017}, 2017.

\bibitem[Weinberger \& Saul(2009)Weinberger and Saul]{lmnn}
Kilian~Q Weinberger and Lawrence~K Saul.
\newblock Distance metric learning for large margin nearest neighbor
  classification.
\newblock \emph{Journal of Machine Learning Research}, 2009.

\bibitem[Wolf et~al.(2019)Wolf, Debut, Sanh, Chaumond, Delangue, Moi, Cistac,
  Rault, Louf, Funtowicz, Davison, Shleifer, von Platen, Ma, Jernite, Plu, Xu,
  Scao, Gugger, Drame, Lhoest, and Rush]{huggingface}
Thomas Wolf, Lysandre Debut, Victor Sanh, Julien Chaumond, Clement Delangue,
  Anthony Moi, Pierric Cistac, Tim Rault, Rémi Louf, Morgan Funtowicz, Joe
  Davison, Sam Shleifer, Patrick von Platen, Clara Ma, Yacine Jernite, Julien
  Plu, Canwen Xu, Teven~Le Scao, Sylvain Gugger, Mariama Drame, Quentin Lhoest,
  and Alexander~M. Rush.
\newblock Huggingface's transformers: State-of-the-art natural language
  processing.
\newblock \emph{ArXiv}, abs/1910.03771, 2019.

\bibitem[Xiao et~al.(2020)Xiao, Zhang, Li, Sun, Tian, Wu, and Wang]{ernie}
Dongling Xiao, Han Zhang, Yukun Li, Yu~Sun, Hao Tian, Hua Wu, and Haifeng Wang.
\newblock Ernie-gen: An enhanced multi-flow pre-training and fine-tuning
  framework for natural language generation.
\newblock \emph{IJCAI}, 2020.

\bibitem[Yang et~al.(2019)Yang, Cheng, Liu, and Sun]{omission-cont}
Zonghan Yang, Yong Cheng, Yang Liu, and Maosong Sun.
\newblock Reducing word omission errors in neural machine translation: A
  contrastive learning approach.
\newblock \emph{Annual Meeting of the Association for Computational
  Linguistics, {ACL}}, 2019.

\bibitem[Yu et~al.(2017)Yu, Zhang, Wang, and Yu]{seqgan}
Lantao Yu, Weinan Zhang, Jun Wang, and Yong Yu.
\newblock Seqgan: sequence generative adversarial nets with policy gradient.
\newblock \emph{Proceedings of the Thirty-First AAAI Conference on Artificial
  Intelligence}, 2017.

\bibitem[Zhang et~al.(2020)Zhang, Zhao, Saleh, and Liu]{pegasus}
Jingqing Zhang, Yao Zhao, Mohammad Saleh, and Peter~J Liu.
\newblock Pegasus: Pre-training with extracted gap-sentences for abstractive
  summarization.
\newblock \emph{ICML}, 2020.

\bibitem[Zhang et~al.(2017)Zhang, Gan, Fan, Chen, Henao, Shen, and
  Carin]{feature-gan}
Yizhe Zhang, Zhe Gan, Kai Fan, Zhi Chen, Ricardo Henao, Dinghan Shen, and
  Lawrence Carin.
\newblock Adversarial feature matching for text generation.
\newblock \emph{International Conference on Machine Learning, {ICML}}, 2017.

\bibitem[Zhang et~al.(2018)Zhang, Galley, Gao, Gan, Li, Brockett, and
  Dolan]{gan-dialogue}
Yizhe Zhang, Michel Galley, Jianfeng Gao, Zhe Gan, Xiujun Li, Chris Brockett,
  and Bill Dolan.
\newblock Generating informative and diverse conversational responses via
  adversarial information maximization.
\newblock \emph{Advances in Neural Information Processing Systems}, 2018.

\bibitem[Zhu et~al.(2019)Zhu, Cheng, Gan, Sun, Goldstein, and Liu]{freelb}
Chen Zhu, Yu~Cheng, Zhe Gan, Siqi Sun, Tom Goldstein, and Jingjing Liu.
\newblock Freelb: Enhanced adversarial training for natural language
  understanding.
\newblock In \emph{International Conference on Learning Representations}, 2019.

\end{thebibliography}
